\pdfoutput=1
\documentclass[11pt]{article}
\usepackage[final]{acl}
\usepackage{times}
\usepackage{latexsym}
\usepackage{amsmath}
\usepackage{amssymb}
\usepackage{graphicx}
\usepackage{tcolorbox}
\tcbuselibrary{skins}
\usepackage{pgfplots}
\usepackage{booktabs}
\usepackage{tabularx}
\usepackage{adjustbox}
\usepackage[table]{xcolor}
\usepackage{pgf}
\usepackage{wrapfig}
\usepackage{arydshln}
\usepackage[T1]{fontenc}
\usepackage[utf8]{inputenc}
\usepackage{microtype}
\usepackage{inconsolata}
\usepackage{multicol}
\usepackage{subcaption}
\usepackage{hyperref}
\usepackage{float}

\title{The Art of Saying ``Maybe'': A Conformal Lens for Uncertainty Benchmarking in VLMs}

\author{
  \textbf{Asif Azad\textsuperscript{1$\ast$$\dagger$}}
  \textbf{Mohammad Sadat Hossain\textsuperscript{1$\ast$}}
  \textbf{MD Sadik Hossain Shanto\textsuperscript{1$\ast$}}
\\
  \textbf{M Saifur Rahman\textsuperscript{1}}
  \textbf{Md Rizwan Parvez\textsuperscript{2}}
\\
\\
  \textsuperscript{1}Bangladesh University of Engineering and Technology \\ 
  \textsuperscript{2}Qatar Computing Research Institute
\\
  \small{\texttt{asifazad0178@gmail.com}, \texttt{sadat@cse.buet.ac.bd}} \\
  \small{\texttt{shantosadikrglhs@gmail.com}, \texttt{mrahman@cse.buet.ac.bd}} \\
  \small{\texttt{mparvez@hbku.edu.qa}}
}

\begin{document}
\maketitle
\renewcommand{\thefootnote}{\fnsymbol{footnote}}
\footnotetext[1]{Equal contribution.}
\footnotetext[2]{Corresponding author.}
\renewcommand{\thefootnote}{\arabic{footnote}}
\begin{abstract}
  Vision-Language Models (VLMs) have achieved remarkable progress in complex visual understanding across scientific and reasoning tasks. While performance benchmarking has advanced our understanding of these capabilities, the critical dimension of uncertainty quantification has received insufficient attention. Therefore, unlike prior conformal prediction studies that focused on limited settings, we conduct a comprehensive uncertainty benchmarking study, evaluating 18 state-of-the-art VLMs (open and closed-source) across 6 multimodal datasets with 3 distinct scoring functions. For closed-source models lacking token-level logprob access, we develop and validate instruction-guided likelihood proxies. Our findings demonstrate that larger models consistently exhibit better uncertainty quantification; models that know more also know better what they don't know. More certain models achieve higher accuracy, while mathematical and reasoning tasks elicit poorer uncertainty performance across all models compared to other domains. This work establishes a foundation for reliable uncertainty evaluation in multimodal systems.
\end{abstract}

\section{Introduction}
\label{sec:intro}

Recent advances in large vision-language models (VLMs) have led to remarkable progress in complex visual understanding and reasoning across diverse domains such as mathematics \citep{wang2024measuring}, science \citep{lu2022learn}, and medicine \citep{WorldMedQA-V2024}. These models now achieve impressive results on challenging multimodal benchmarks, demonstrating their potential for real-world impact.

Yet, despite these capabilities, significant challenges remain. As VLMs are increasingly deployed in high-stakes domains like medical diagnostics \citep{li2025vision}, educational assessments, and scientific reasoning, the consequences of model failure become critical. While accuracy metrics highlight overall performance, they do not reveal when a model is uncertain or likely to err. In practical applications, especially in sensitive fields like healthcare, an overconfident but incorrect prediction can have severe repercussions. Thus, quantifying and understanding model uncertainty in a computationally efficient way is essential for building reliable and trustworthy VLM systems.

Quantifying uncertainty in VLMs is therefore crucial for building reliable and trustworthy systems, especially in high-stakes domains. While classical approaches such as Bayesian neural networks \citep{blundell2015weight}, deep ensembles \citep{lakshminarayanan2017simple}, and calibration-based methods \citep{guo2017calibration} have been explored for uncertainty estimation mostly in traditional machine learning models, their application to foundation models (e.g., LLMs, VLMs, and multimodal architectures), where parameters often scale to billions or trillions, is limited by computational cost and scalability issues. Conformal prediction, in contrast, offers a computationally feasible, model-agnostic framework with formal statistical guarantees, making it particularly attractive for uncertainty quantification in complex multimodal settings. Prior work has applied conformal prediction to LLMs for benchmarking predictive confidence \citep{ye2024benchmarking}, but its utility for VLMs, where uncertainty arises from both visual and textual modalities, remains largely unexplored. This motivates our study, which systematically investigates conformal prediction as a principled approach for uncertainty benchmarking in VLMs across diverse set of tasks.

This study is guided by several core research questions:
\begin{enumerate}
  \item Do different conformal scoring functions yield similar efficiency in terms of prediction set size, or do their behaviors diverge across tasks and models?
  \item Is there a correlation between model accuracy and the size of conformal prediction sets, indicating calibration quality?
  \item How do uncertainty metrics (set size) vary with model scale and architecture?
  \item Can this uncertainty quantification approach be applied to black-box proprietary models, provided they expose token-level probabilities?
  \item How does uncertainty calibration vary across domains and do certain model families show systematic advantages?
\end{enumerate}

Our evaluation spans a suite of carefully chosen datasets- MMMU~\citep{yue2024mmmu}, MMMU-Pro~\citep{yue2024mmmupro}, AI2D~\citep{kembhavi2016diagram}, MathVision~\citep{wang2024measuring,wang2025mathcodervl}, ScienceQA~\citep{lu2022learn}, and WorldMedQAV~\citep{WorldMedQA-V2024}- each probing distinct aspects of visual and scientific understanding. We systematically compare multiple scoring functions within the conformal framework to provide a comprehensive analysis of uncertainty in VLMs.

Our findings reveal patterns of uncertainty that correlate not only with accuracy but also with task modality and semantic complexity, offering deeper insights into when and why VLMs hesitate.

\section{Related Works}
Uncertainty quantification has long been central to machine learning, especially for risk-sensitive decisions \citep{Abdar_2021}. Classic methods include Bayesian neural networks \citep{blundell2015weight}, deep ensembles \citep{lakshminarayanan2017simple}, and calibration techniques \citep{guo2017calibration}. These work well in low dimensions but can be computationally heavy or lack expressiveness for deep multimodal models.

Conformal prediction provides robust, statistical guarantees and applies across domains \citep{zhou2025conformalpredictiondataperspective}. It's distribution-free, model-agnostic, and efficient, making it ideal for large-scale models. Recent efforts have used it for LLMs \citep{angelopoulos2021gentle,ye2024benchmarking} and VLMs \citep{kostumov2024uncertaintyawareevaluationvisionlanguagemodels}, delivering coverage via prediction sets. Still, past work focused on text-only models or simpler benchmarks with outdated VLMs. In contrast, while \citet{kostumov2024uncertaintyawareevaluationvisionlanguagemodels} mainly examined accuracy–uncertainty alignment and scaling effects, our study investigates a broader set of five research questions covering scoring efficiency, calibration–accuracy correlations, architecture and scale influences, black-box model applicability, and domain-specific trends. We also evaluate a larger and more diverse suite of state-of-the-art proprietary and open VLMs across multimodal reasoning tasks, extending beyond alignment analysis toward a unified, theoretically grounded uncertainty benchmarking framework.

VLMs have spotlighted multimodal understanding, with benchmarks targeting visual reasoning \citep{zellers2019recognitioncognitionvisualcommonsense}, hallucination detection \citep{liu2022tokenlevelreferencefreehallucinationdetection, sadat-etal-2023-delucionqa, islam-etal-2024-open}, and multimodal knowledge \citep{xu2023lvlmehubcomprehensiveevaluationbenchmark}.

Related research covers hallucination \citep{rawte2023surveyhallucinationlargefoundation}, interpretability \citep{bommasani2022opportunitiesrisksfoundationmodels}, and confidence estimation \citep{hendrycks2018baselinedetectingmisclassifiedoutofdistribution}. These offer insights but often miss formal guarantees for uncertainty quantification. Our approach uses conformal prediction for a unified, distribution-free framework that's practical for big foundation models and backed by theory. It outperforms heuristic hallucination detectors or post-hoc interpretability methods, paving a solid way to benchmark VLM uncertainty.

\section{Conformal Prediction}
Conformal prediction offers a statistically rigorous, distribution-free framework for uncertainty quantification. It builds prediction sets that include the true output with a specified probability. For any model \( f \) mapping input \( X \) to a probability distribution over finite label space \( Y \), it creates \( C(X) \subseteq Y \) such that:

\begin{equation}
  \mathbb{P}(Y_{\text{true}} \in C(X)) \geq 1 - \alpha,
\end{equation}
where \( \alpha \) is the error rate.

To build these sets, use a score function \( s(X, y) \) that measures incompatibility between \( X \) and \( y \). The process is:

\begin{enumerate}
  \item Compute scores \( s_i = s(X_i^{\text{cal}}, Y_i^{\text{cal}}) \) for calibration set \( D_{\text{cal}} = \{(X_1^{\text{cal}}, Y_1^{\text{cal}}), \ldots, (X_n^{\text{cal}}, Y_n^{\text{cal}})\} \).
  \item Find threshold \( \hat{q} \) as the \( \lceil(n+1)(1-\alpha)\rceil/n \) quantile of scores:
        \begin{equation}
          \hat{q} = \text{quant}\bigl(\{s_1, \ldots, s_n\}, \lceil(n+1)(1-\alpha)\rceil/n\bigr)
        \end{equation}
  \item For test input \( X \), form \( C(X) = \{y \in \mathcal{Y} : s(X, y) \leq \hat{q}\} \):
        \begin{equation}
          C(X) = \{y \in \mathcal{Y} : s(X, y) \leq \hat{q}\}
        \end{equation}
\end{enumerate}

In practice, it uses heuristic scoring functions, which influence calibration and set size. We examine three: Least Ambiguous Classifier (LAC), Adaptive Prediction Sets (APS), and Marginal Score (MS), comparing them across tasks and models. We also check entropy-based uncertainty as a complement.

Common scoring functions for classification include:

\paragraph{Least Ambiguous Classifier (LAC).} The LAC score \citep{sadinle2019least} is
\begin{equation}
  s_{\text{LAC}}(X, y) = 1 - f(X)_y,
\end{equation}
where \( f(X)_y \) is the predicted probability for \( y \). It penalizes low-confidence predictions, favoring confident ones in sets.

\paragraph{Adaptive Prediction Sets (APS).} The APS score \cite{romano2020classification} is
\begin{equation}
  s_{\text{APS}}(X, y) = \sum_{y' : f(X)_{y'} \geq f(X)_y} f(X)_{y'},
\end{equation}
summing probabilities of classes as or more supported than \( y \). It adjusts set size to prediction ambiguity, suiting diffuse distributions.

\paragraph{Marginal Score.} The margin score is
\begin{equation}
  s_{\text{margin}}(X, y) = f(X)_{(1)} - f(X)_{(2)},
\end{equation}
where \( f(X)_{(1)} \) and \( f(X)_{(2)} \) are top-1 and top-2 probabilities. It measures confidence gap, ideal for high-ambiguity tasks.

\section{Datasets}
We test VLM uncertainty on six diverse, challenging datasets probing multimodal reasoning:

\paragraph{MMMU} The Massive Multi-discipline Multimodal Understanding (MMMU) dataset~\citep{yue2024mmmu} assesses VLMs on college-level questions across 30 disciplines like science, medicine, engineering, and humanities.

\paragraph{MMMU-Pro} MMMU-Pro \citep{yue2024mmmupro} extends MMMU with tougher, professional questions emphasizing real-world scenarios and domain expertise, ramping up visual and textual complexity.

\paragraph{ScienceQA} ScienceQA~\citep{lu2022learn} focuses on elementary/middle school science with over 21,000 questions in natural sciences, physics, and biology, often with images like diagrams. It tests visual-scientific integration.

\paragraph{AI2D} The AI2 Diagrams (AI2D) dataset~\citep{kembhavi2016diagram} has over 15,000 elementary science questions paired with labeled diagrams and multiple-choice answers, demanding interpretation of visuals, spatial relations, and concepts.

\paragraph{MathVision} MathVision~\citep{wang2024measuring,wang2025mathcodervl} is a visual math benchmark with problems in images like graphs or equations, evaluating extraction of quantitative info and math reasoning.

\paragraph{WorldMedQAV} WorldMedQAV \citep{WorldMedQA-V2024} features clinical images (e.g., X-rays, slides) with expert multiple-choice questions, testing medical image interpretation and diagnostic reasoning in healthcare contexts.

These datasets create a broad testbed for VLM uncertainty across domains, modalities, and reasoning types.

\section{Experimentation}

\subsection{Prompting}
We used a three-part prompting strategy across all datasets. First, dataset-specific system messages set the VLM's role, like ``scientific diagram analyzer'' for AI2D or ``medical image diagnostician'' for WorldMedQAV. These guided models to the domain while keeping instructions consistent.

Second, zero-shot task instructions gave a quick overview of the question type, without hints on solving. For example, MathVision prompts started with ``I will show you an image along with a multiple-choice math question.'' This setup offered context but avoided biasing responses.

Finally, every prompt ended with ``Only respond with the option letter'' to standardize outputs for uncertainty analysis. This kept prompt differences from skewing results. Full prompts are in Appendix~\ref{appendix:prompts}.

\begin{table*}[htbp]
  \centering
  \includegraphics[width=\linewidth]{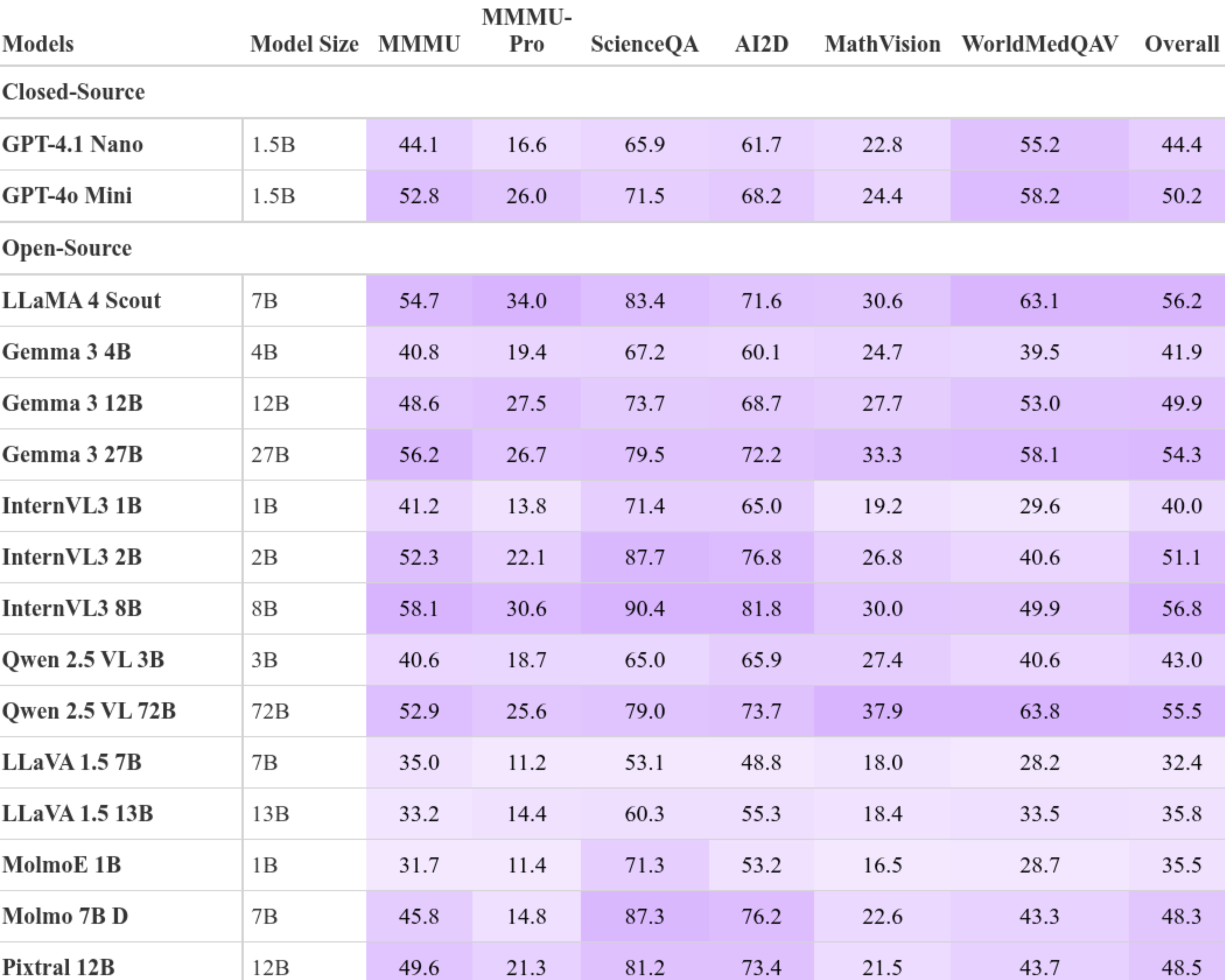}
  \caption{Accuracy ($\uparrow$) performance (\%) of VLMs across six benchmarking datasets. Color intensity indicates higher performance.}
  \label{fig:acc-table}
\end{table*}

\begin{table*}[htbp]
  \centering
  \includegraphics[width=\textwidth]{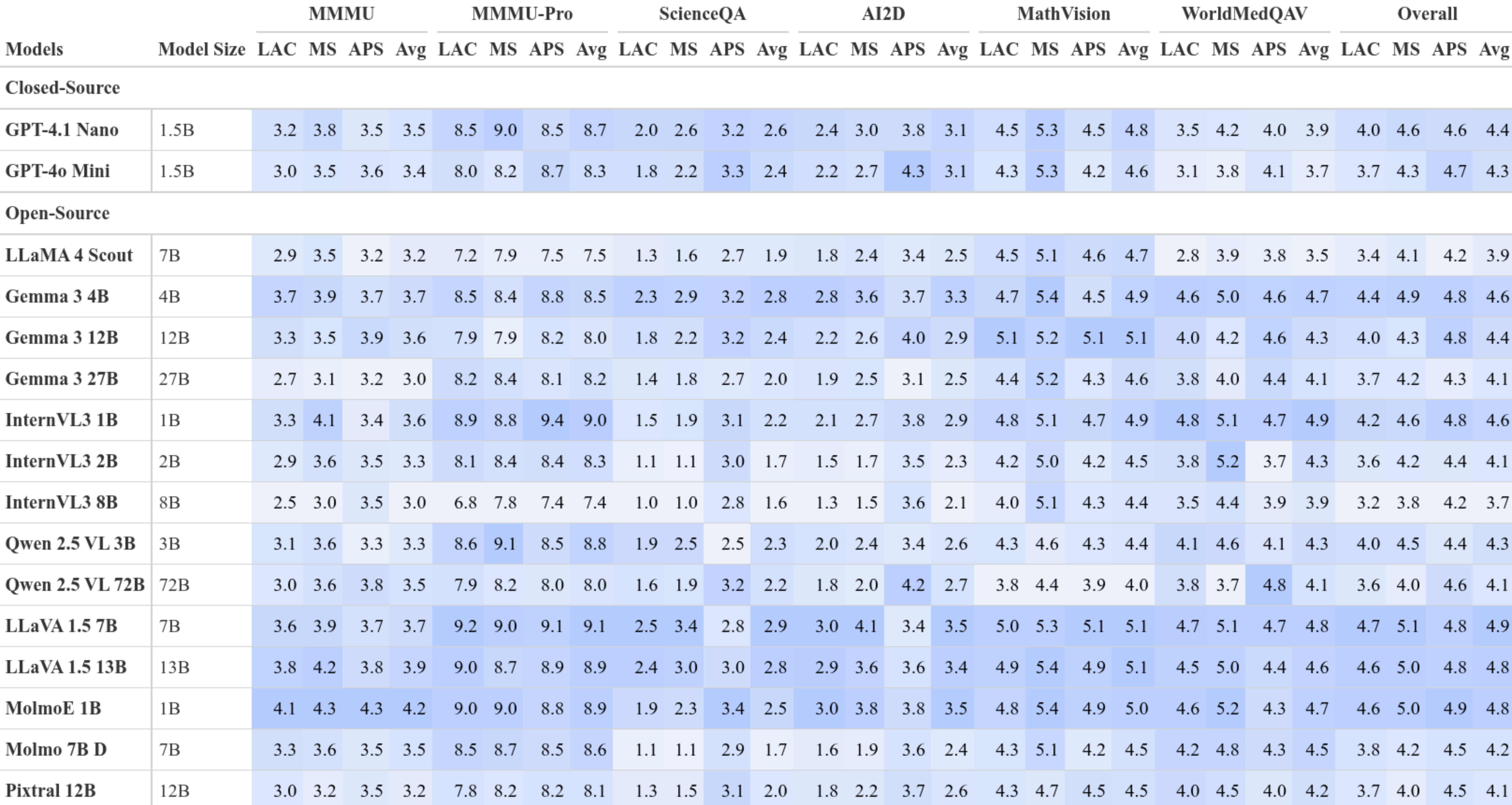}
  \caption{Set Size ($\downarrow$) results across models, datasets, and conformal scoring functions (LAC, MS, and APS). Lower values indicate more precise uncertainty quantification.}
  \label{fig:ss-table}
\end{table*}

\subsection{Inference Setup}
We ran a controlled inference pipeline, adapting to model size and availability. Small models ($\leq$ 7B parameters) used P100 and T4 GPUs on Kaggle.

\begin{figure*}[htb!]
  \centering
  \includegraphics[width=0.9\linewidth]{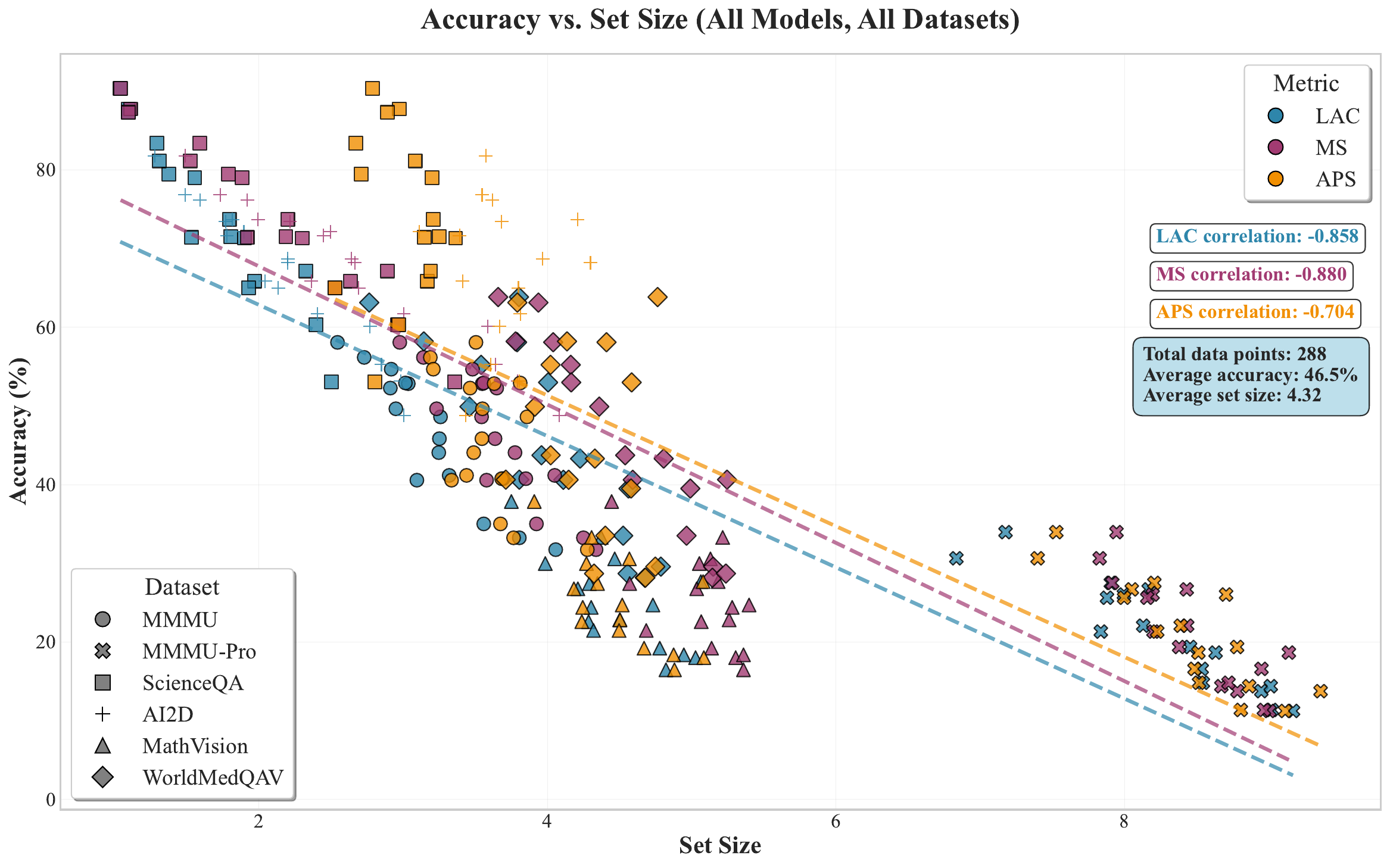}
  \caption{Correlation between accuracy and set size across datasets. Higher-performing models produce more concentrated prediction sets.}
  \label{fig:ac-ss}
\end{figure*}

For VLMs, we preferred OpenRouter's API when available, covering large and mid-sized models. For mid-sized ones without API, we turned to A1000 GPUs on Runpod for smooth runs.

\begin{figure*}[htb!]
  \centering
  \includegraphics[width=0.86\textwidth]{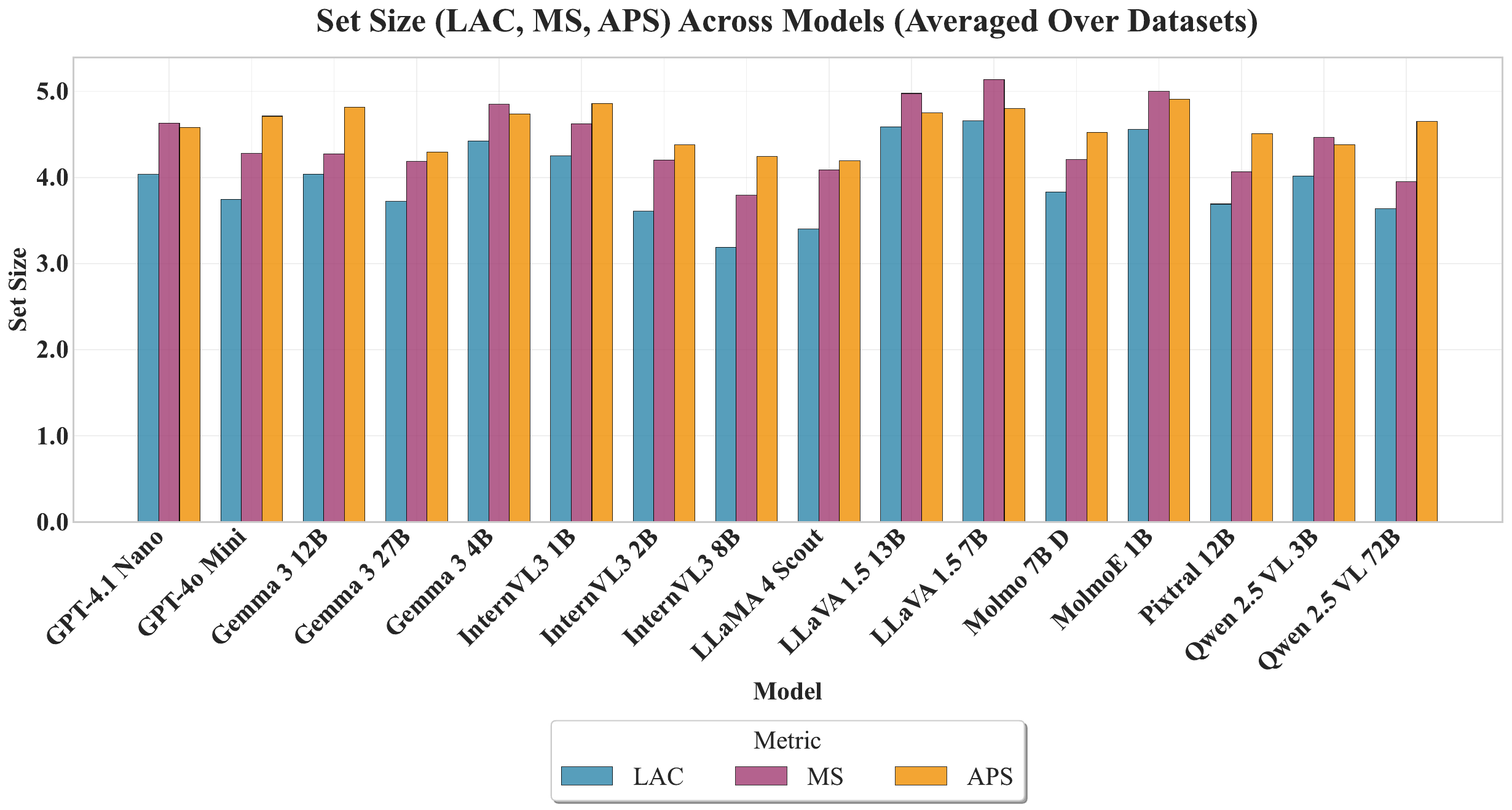}
  \caption{Comparison of set sizes across VLMs and scoring functions. LAC scoring consistently produces the most compact prediction sets.}
  \label{fig:ss-models}
\end{figure*}

Models on Kaggle and Runpod came straight from Hugging Face repositories to use official versions. Across all, we set \texttt{do\_sample=False} for greedy decoding, skipping temperature, top-k, and top-p.

From each response, we pulled log-probabilities for answer letters (A, B, C, D) via token-level scores on the final output. With multiple-choice tasks limited to one letter, we grabbed the predicted answer's log-prob. These distributions drove our conformal prediction at $\alpha = 0.1$ (90\% coverage), splitting datasets 50/50 for calibration and testing.

\subsection{Evaluated Models}
We tested 18 vision-language models spanning architectures and scales. Open-source picks include Llama-4-Scout, Gemma-3 \citep{gemmateam2025gemma3technicalreport} (4B/12B/27B), InternVL3 \citep{zhu2025internvl3exploringadvancedtraining} (1B/2B/8B), Molmo variants \citep{deitke2024molmopixmoopenweights} (1B/7B), Qwen2.5-VL \citep{bai2025qwen25vltechnicalreport} (3B/72B), Llava-1.5 \citep{liu2024improvedbaselinesvisualinstruction} (7B/13B), and Pixtral \citep{agrawal2024pixtral12b} (12B). Proprietary ones include GPT-4.1-nano and GPT-4o-mini, which were the only commercial VLMs sharing token probabilities for conformal work. Later we extended our work with proxy methods, to include Gemini-2.5-Flash and Claude Haiku-4.5.

This lineup lets us compare uncertainty traits across sizes, builds, and open/closed paradigms. Since models like Gemini and Claude do not expose token-level probabilities through their APIs, we developed an instruction-guided likelihood proxy (Section~\ref{sec:proxy-methods}) to extend our evaluation to these closed-source VLMs.

\subsection{Evaluation Metrics}
Our main UQ metric is \textit{Set Size} (SS), averaging conformal set sizes:

\begin{equation}
  \text{SS} = \frac{1}{|D_{test}|}\sum_{(X_t,Y_t)\in D_{test}} |C(X_t)|
\end{equation}

Lower SS means sharper uncertainty, with SS=1 for perfect certainty on correct predictions. We pair it with \textit{Accuracy} (Acc) for prediction performance:

\begin{equation}
  \text{Acc} = \frac{1}{|D_{test}|}\sum_{(X_t,Y_t)\in D_{test}} \mathbb{I}(Y_p = Y_t)
\end{equation}

We also check \textit{Coverage Rate} (CR) to confirm guarantees:

\begin{equation}
  \text{CR} = \frac{1}{|D_{test}|}\sum_{(X_t,Y_t)\in D_{test}} \mathbb{I}(Y_t \in C(X_t))
\end{equation}

CR should hit at least $(1-\alpha)$ overall. These metrics (SS, Acc, CR) across LAC, MS, and APS give a full picture of prediction quality and uncertainty trustworthiness.

\section{Results}

\begin{table*}[htb]
  \centering
  \includegraphics[width=\textwidth]{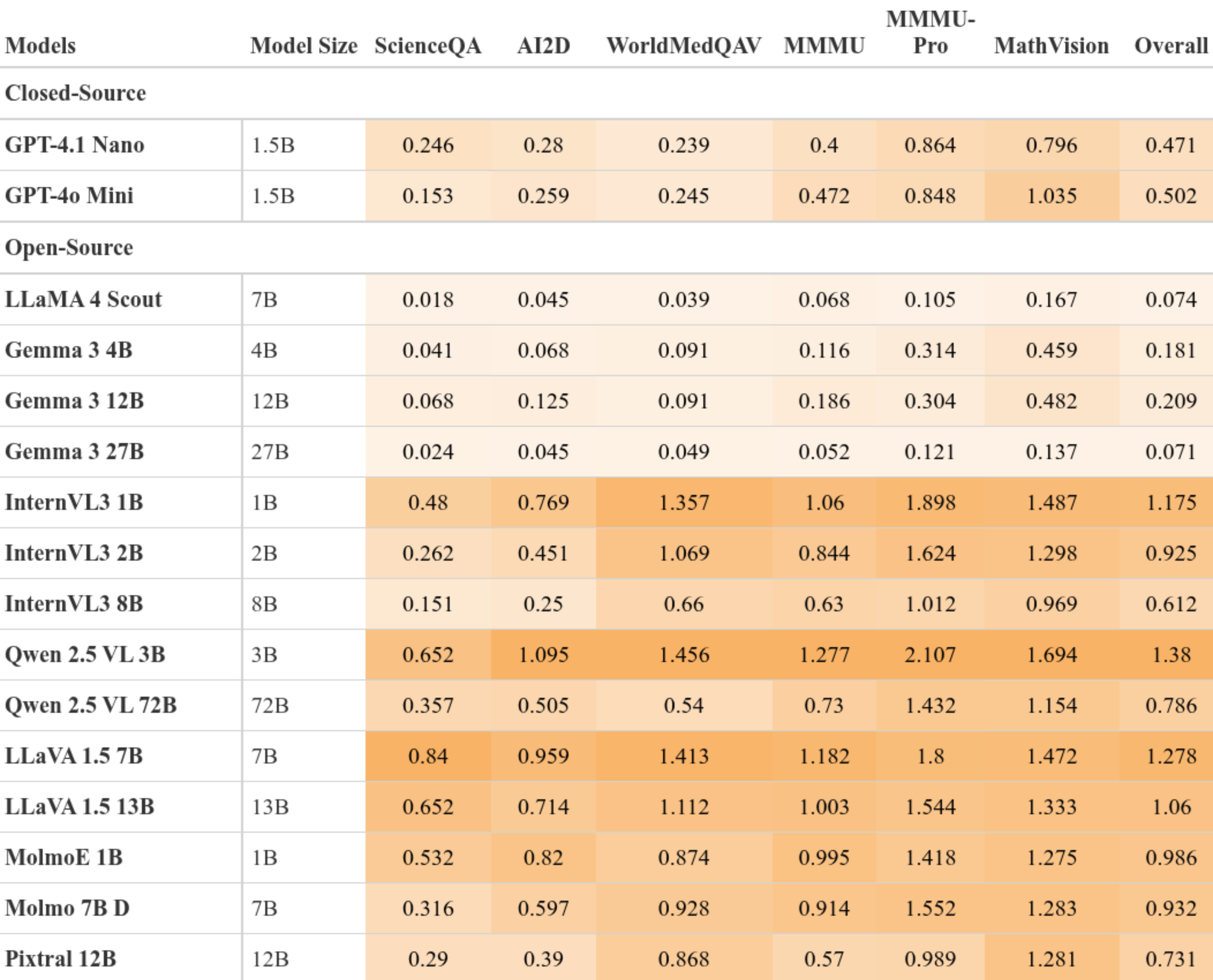}
  \caption{Entropy ($\downarrow$) based uncertainty scores across models and datasets. Lower values indicate better uncertainty quantification.}
  \label{fig:entropy-table}
\end{table*}

\subsection{Uncertainty Performance Analysis}

Table~\ref{fig:ss-table} presents set sizes for our VLMs and conformal scoring functions. LAC yields the smallest sets for most models, highlighting its strength in vision-language uncertainty quantification. Larger models in each family (e.g., Qwen-VL 72B vs. 3B) show consistently smaller sets, underscoring scaling's role in better calibration.

Figure~\ref{fig:ac-ss} illustrates the strong inverse link between accuracy and set size: top performers create tighter, better-calibrated sets. As detailed in Appendix~\ref{sec:add-results}, Figure~\ref{fig:all-relations}, this negative correlation is a key principle, stronger models reflect appropriate confidence.

Set size distributions across methods appear in Appendix Figure~\ref{fig:ss-bar-chart-appendix}, where APS shows the lowest variability.

\subsection{Accuracy Performance Analysis}

Table~\ref{fig:acc-table} highlights key trends: (1) larger family members hit higher accuracy; (2) datasets vary widely, with MMMU-Pro the toughest (20.85\% average) and ScienceQA the easiest (73.78\%). Among open-source models, InternVL 8B leads, shining on AI2D and ScienceQA.

\begin{figure*}[ht!]
  \centering
  \includegraphics[width=\textwidth]{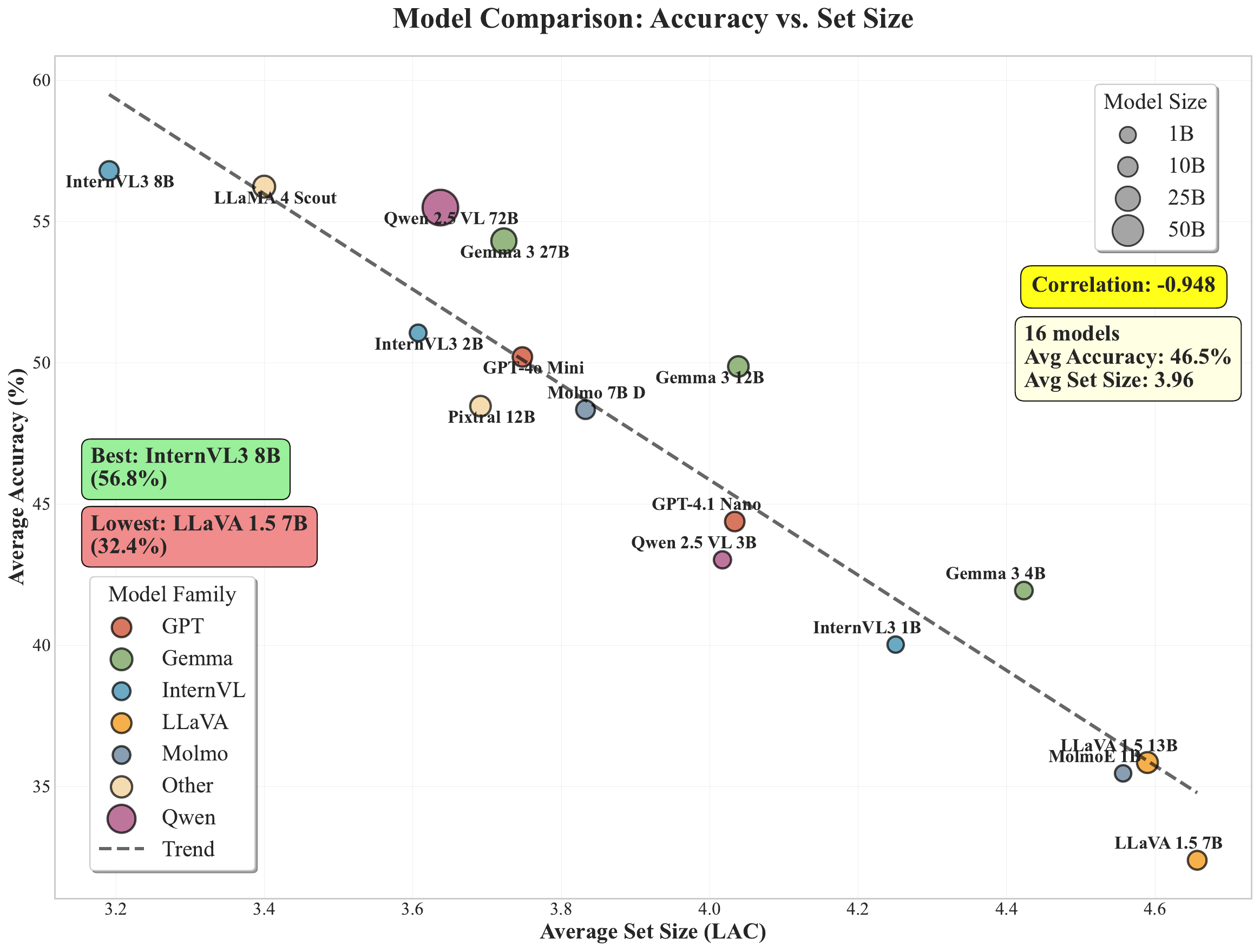}
  \caption{Relationship between model size, accuracy, and set size. Larger models exhibit both higher accuracy and smaller set sizes.}
  \label{fig:model-size-ac-ss}
\end{figure*}

Figure~\ref{fig:model-size-ac-ss} confirms model size boosts accuracy and shrinks set sizes. Larger models (>10B) cluster in the high-accuracy, low-set-size zone, proving scaling enhances performance and calibration.

\subsection{Coverage Rate Analysis}

Coverage validation is in Appendix Table~\ref{fig:cr-table-appendix} and Figure~\ref{fig:cr-models-ds-comp-appendix}. Our framework meets $(1-\alpha)=90\%$ coverage in most cases, with minor slips. It's toughest on complex tasks like MathVision and MMMU-Pro, yet holds up across domains.

\subsection{Model-Specific Uncertainty Performance}

\begin{figure*}[ht!]
  \centering
  \includegraphics[width=\textwidth]{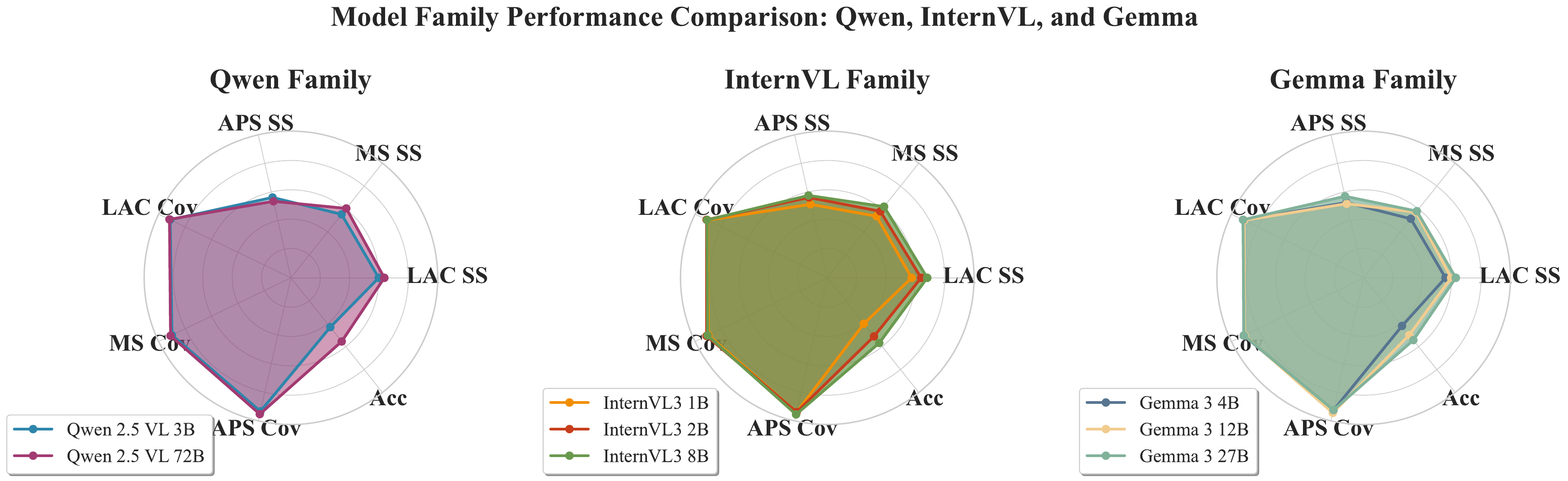}
  \caption{Uncertainty profiles for three model families - InternVL (1B, 2B, 8B), Qwen-VL (3B, 72B), and Gemma (4B, 12B, 27B). Smaller enclosed radar areas indicate better-calibrated uncertainty. Darker shades represent larger models within each family. Each family exhibits distinct scaling patterns across domains. Metrics include Accuracy (Acc.), Coverage for LAC/MS/APS (\emph{higher values closer to 90\% are better}), and Inverted Set Size for LAC/MS/APS (\emph{smaller set sizes are better, inverted for visualization}).}
  \label{fig:model-family-radars}
\end{figure*}

Figure~\ref{fig:model-family-radars} uncovers unique ``uncertainty signatures'' per family. Across Gemma, Qwen-VL, and InternVL, bigger variants show superior confidence calibration at the target coverage.

\begin{figure}[ht!]
  \centering
  \includegraphics[width=\linewidth]{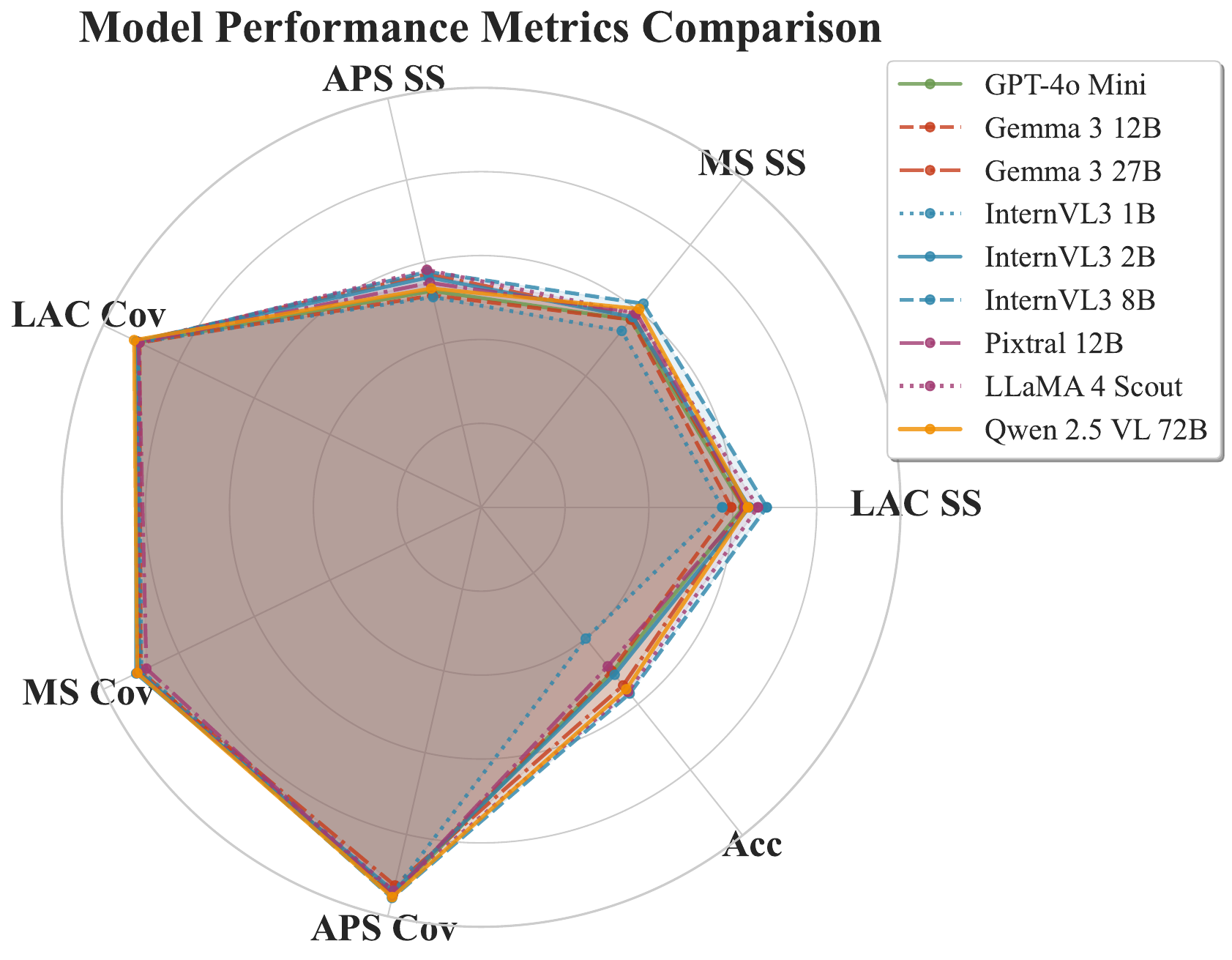}
  \caption{Comparative uncertainty profiles across all VLMs. Proprietary models like GPT-4o-mini achieve remarkably well-calibrated uncertainty estimates.}
  \label{fig:radar-all}
\end{figure}

Figure~\ref{fig:radar-all} compares all families: InternVL tops uncertainty quantification, outpacing rivals. Llama-4-Scout ranks second, hinting at architectural edges for calibration in these lines.

\subsection{Domain-Specific Uncertainty Performance}

VLMs calibrate uncertainty better when visuals support reasoning, not lead it. ScienceQA, where images back text, scores high accuracy (75.2\% average) and tight sets (2.1 average). MathVision, needing exact visual number pulls, challenges calibration (4.5 average set size, lower accuracy).

In medical tasks (WorldMedQAV), calibration stays steady within size groups despite accuracy swings. This implies domain visual skills and uncertainty awareness evolve separately, models might spot features without gauging confidence accurately, or the reverse.

\subsection{Entropy-Based Uncertainty Analysis}

Table~\ref{fig:entropy-table} shows entropy scores, mirroring conformal results: larger models demonstrate lower entropy (better calibration), with the Gemma 12B vs. 4B outlier in set sizes. Gemma leads overall, with all sizes in top spots. Domains follow suit; ScienceQA exhibit the lowest entropy, and MathVision the highest. This backs our main findings and extends them to alternative uncertainty methods.

\begin{table*}[htbp!]
  \centering
  \includegraphics[width=\textwidth]{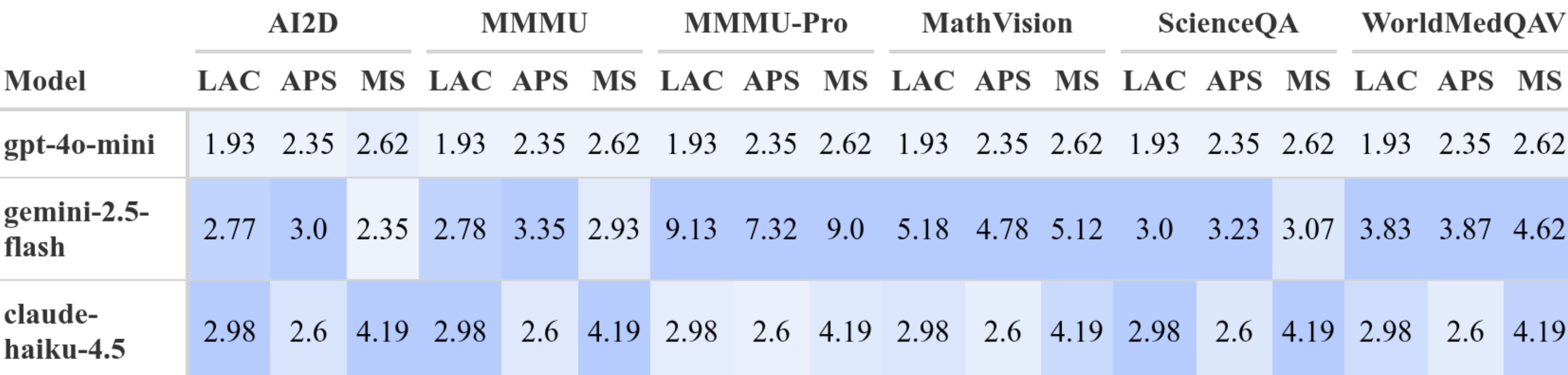}
  \caption{Set Size ($\downarrow$) for closed-source VLMs using the instruction-guided likelihood proxy.}
  \label{fig:closed-ss-table}
\end{table*}

\begin{table*}[htbp!]
  \centering
  \resizebox{\textwidth}{!}{
    \begin{tabular}{l l l p{7.5cm}}
      \toprule
      \textbf{Task Characteristic}       & \textbf{Datasets} & \textbf{Scoring} & \textbf{Adaptation Rationale}                     \\
      \midrule
      Low Ambiguity / Fact-Based         &
      ScienceQA, AI2D                    &
      LAC                                &
      Probability mass is concentrated; LAC yields the smallest sets while maintaining 90\% coverage, reducing unnecessary options. \\
      \midrule
      High Ambiguity / Complex Reasoning &
      MathVision, MMMU-Pro               &
      APS                                &
      Distributions are diffuse; APS accumulates probability mass adaptively, improving robustness under reasoning uncertainty.     \\
      \midrule
      Differential / Binary Ambiguity    &
      WorldMedQAV                        &
      MS                                 &
      Tasks often involve two leading hypotheses; MS models confidence gaps, avoiding low-relevance tail options.                   \\
      \bottomrule
    \end{tabular}
  }
  \caption{Task-adaptive scoring function recommendations based on task ambiguity and uncertainty structure.}
  \label{fig:task-matching}
\end{table*}

\subsection{Instruction-Guided Uncertainty Proxies}
\label{sec:proxy-methods}
\begin{table}[ht!]
  \centering
  \includegraphics[width=\linewidth]{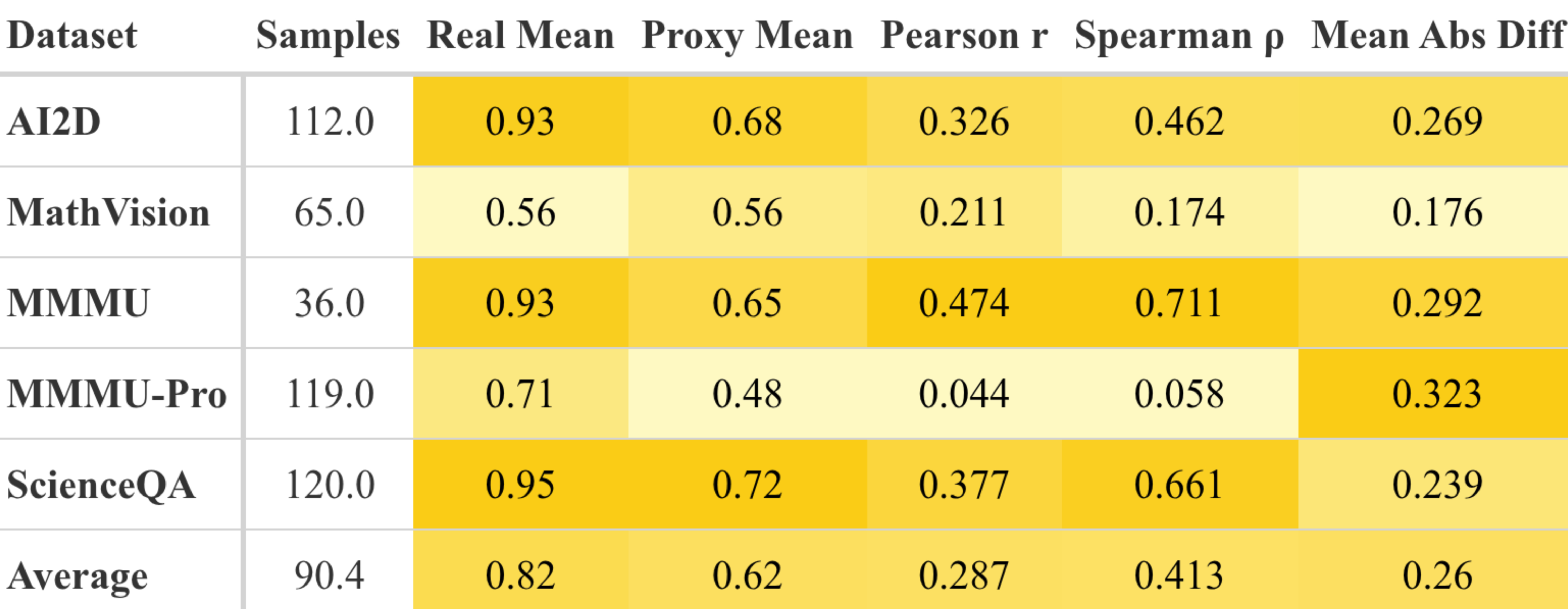}
  \caption{Alignment between GPT-4o-mini's true logprobs and instruction-guided proxy likelihoods across six benchmarks. Moderate-to-strong correlation validates proxy-based uncertainty for closed-source VLMs.}
  \label{fig:proxy-alignment}
\end{table}

To evaluate closed-source VLMs without logprob access, we adopt an instruction-guided likelihood proxy where models output option-wise numerical likelihoods in a strict JSON format, used as an uncertainty signal.

We validate this proxy on GPT-4o-mini by comparing it against true token-level logprobs. Table~\ref{fig:proxy-alignment} shows moderate-to-strong alignment (Pearson up to 0.47, Spearman up to 0.71, $\sim$74\% directional agreement), indicating that the proxy preserves relative uncertainty structure.

Using this proxy, we evaluate GPT-4o-mini, Gemini-2.5-Flash, and Claude Haiku~4.5. Set-size results in Table~\ref{fig:closed-ss-table} mirror earlier findings: stronger models yield smaller, better-calibrated sets, while visually intensive datasets (MathVision, MMMU-Pro) produce larger sets for weaker models. Accuracy and coverage trends remain unchanged (Appendix Tables~\ref{fig:closed-cr-appendix}, \ref{fig:closed-acc-appendix}).

\subsection{Task-Adaptive Scoring Functions}

We study how conformal scoring functions vary with task characteristics and derive lightweight selection guidelines. Using prediction entropy as a proxy for task ambiguity (Table~\ref{fig:entropy-table}), we link ambiguity with set efficiency and coverage behavior (Tables~\ref{fig:ss-table},~\ref{fig:cr-table-appendix}).

Table~\ref{fig:task-matching} summarizes consistent patterns: LAC performs best in low-ambiguity, fact-based tasks; APS is more robust in high-ambiguity reasoning domains; and MS is effective for differential or near-binary uncertainty. These results suggest that task-aware scoring selection yields better-calibrated uncertainty than a uniform choice.

\section{Conclusion}
This work presents a comprehensive conformal uncertainty benchmarking framework for Vision-Language Models (VLMs), systematically analyzing how uncertainty behaves across model scales, architectures, and domains. Using conformal prediction, a distribution-free, model-agnostic method with formal guarantees, we evaluated 18 state-of-the-art VLMs over six multimodal benchmarks and three scoring functions.

Our results reveal clear trends: larger models not only achieve higher accuracy but also produce more calibrated uncertainty estimates; they know better what they do not know. Scaling enhances both performance and confidence reliability across model families such as Qwen-VL, Gemma, and InternVL.

Overall, this study establishes a principled foundation for multimodal uncertainty evaluation and highlights conformal prediction as a practical, theoretically grounded tool for measuring trust in large VLMs. By quantifying when models are unsure, our approach moves toward building safer, more accountable, and trustworthy AI systems, enabling future research on dynamic calibration and adaptive uncertainty in real-world applications.

\section*{Limitations}

This work offers key insights on VLM uncertainty, but limitations persist. First, we stuck to multiple-choice datasets, as conformal methods need defined prediction sets amid compute limits. Future efforts could adapt for generative tasks with unbounded outputs, like uncertainty in open text via specialized techniques.

Second, while we expanded our proprietary model evaluation beyond GPT to include Gemini-2.5-Flash and Claude Haiku~4.5, this extension required an instruction-guided likelihood proxy rather than direct token-level logprobs. Most commercial VLM providers - including Google, Anthropic, and others - do not expose token-level log probabilities through their APIs, a design choice that fundamentally limits principled uncertainty quantification. Our proxy method, which prompts models to self-report option-wise likelihoods in a structured format, demonstrates moderate-to-strong alignment with true logprobs (Section~\ref{sec:proxy-methods}) but remains an indirect approximation. This proxy inevitably introduces noise from the model's self-assessment biases, instruction-following variability, and the inherent gap between generated likelihood estimates and actual next-token distributions. If providers were to expose token-level logprobs - as GPT models partially do - our conformal framework could be applied directly, yielding more precise and theoretically grounded uncertainty quantification. Until such interfaces become standard, uncertainty benchmarking for closed-source VLMs will rely on proxy methods that, while practically useful, lack the formal guarantees of direct logprob-based conformal prediction.

\bibliography{custom}

@inproceedings{islam-etal-2024-open,
    title = "Open-{RAG}: Enhanced Retrieval Augmented Reasoning with Open-Source Large Language Models",
    author = "Islam, Shayekh Bin  and
      Rahman, Md Asib  and
      Hossain, K S M Tozammel  and
      Hoque, Enamul  and
      Joty, Shafiq  and
      Parvez, Md Rizwan",
    editor = "Al-Onaizan, Yaser  and
      Bansal, Mohit  and
      Chen, Yun-Nung",
    booktitle = "Findings of the Association for Computational Linguistics: EMNLP 2024",
    month = nov,
    year = "2024",
    address = "Miami, Florida, USA",
    publisher = "Association for Computational Linguistics",
    url = "https://aclanthology.org/2024.findings-emnlp.831/",
    doi = "10.18653/v1/2024.findings-emnlp.831",
    pages = "14231--14244"
}

@inproceedings{sadat-etal-2023-delucionqa,
    title = "{D}elucion{QA}: Detecting Hallucinations in Domain-specific Question Answering",
    author = "Sadat, Mobashir  and
      Zhou, Zhengyu  and
      Lange, Lukas  and
      Araki, Jun  and
      Gundroo, Arsalan  and
      Wang, Bingqing  and
      Menon, Rakesh  and
      Parvez, Md  and
      Feng, Zhe",
    editor = "Bouamor, Houda  and
      Pino, Juan  and
      Bali, Kalika",
    booktitle = "Findings of the Association for Computational Linguistics: EMNLP 2023",
    month = dec,
    year = "2023",
    address = "Singapore",
    publisher = "Association for Computational Linguistics",
    url = "https://aclanthology.org/2023.findings-emnlp.59/",
    doi = "10.18653/v1/2023.findings-emnlp.59",
    pages = "822--835"
}

@article{ye2024benchmarking,
  title={Benchmarking {LLM}s via Uncertainty Quantification},
  author={Ye, Fanghua and Yang, Mingming and Pang, Jianhui and Wang, Longyue and Wong, Derek and Yilmaz, Emine and Shi, Shuming and Tu, Zhaopeng},
  journal={Advances in Neural Information Processing Systems},
  volume={37},
  pages={15356--15385},
  year={2024}
}

@inproceedings{blundell2015weight,
  title={Weight Uncertainty in Neural Network},
  author={Blundell, Charles and Cornebise, Julien and Kavukcuoglu, Koray and Wierstra, Daan},
  booktitle={International conference on machine learning},
  pages={1613--1622},
  year={2015},
  organization={PMLR}
}

@article{lakshminarayanan2017simple,
  title={Simple and Scalable Predictive Uncertainty Estimation Using Deep Ensembles},
  author={Lakshminarayanan, Balaji and Pritzel, Alexander and Blundell, Charles},
  journal={Advances in neural information processing systems},
  volume={30},
  year={2017}
}

@inproceedings{guo2017calibration,
  title={On Calibration of Modern Neural Networks},
  author={Guo, Chuan and Pleiss, Geoff and Sun, Yu and Weinberger, Kilian Q},
  booktitle={International conference on machine learning},
  pages={1321--1330},
  year={2017},
  organization={PMLR}
}

@article{angelopoulos2021gentle,
  title={A Gentle Introduction to Conformal Prediction and Distribution-free Uncertainty Quantification},
  author={Angelopoulos, Anastasios N and Bates, Stephen},
  journal={arXiv preprint arXiv:2107.07511},
  year={2021}
}

@misc{kembhavi2016diagram,
  title        = {A Diagram Is Worth A Dozen Images},
  author       = {Kembhavi, Aniruddha and Salvato, Mike and Kolve, Eric and Seo, Minjoon and Hajishirzi, Hannaneh and Farhadi, Ali},
  year         = {2016},
  eprint       = {1603.07396},
  archivePrefix= {arXiv},
  primaryClass = {cs.CV}
}

@inproceedings{lu2022learn,
    title={Learn to Explain: Multimodal Reasoning via Thought Chains for Science Question Answering},
    author={Lu, Pan and Mishra, Swaroop and Xia, Tony and Qiu, Liang and Chang, Kai-Wei and Zhu, Song-Chun and Tafjord, Oyvind and Clark, Peter and Ashwin Kalyan},
    booktitle={The 36th Conference on Neural Information Processing Systems (NeurIPS)},
    year={2022}
}

@article{sadinle2019least,
  title     = {Least Ambiguous Set-valued Classifiers with Bounded Error Levels},
  author    = {Sadinle, Mauricio and Lei, Jing and Wasserman, Larry},
  journal   = {Journal of the American Statistical Association},
  volume    = {114},
  number    = {525},
  pages     = {223--234},
  year      = {2019},
  publisher = {Taylor \& Francis}
}

@article{romano2020classification,
  title   = {Classification with Valid and Adaptive Coverage},
  author  = {Romano, Yaniv and Sesia, Matteo and Candes, Emmanuel},
  journal = {Advances in neural information processing systems},
  volume  = {33},
  pages   = {3581--3591},
  year    = {2020}
}

@inproceedings{yue2024mmmu,
  title     = {{MMMU}: A Massive Multi-discipline Multimodal Understanding and Reasoning Benchmark for Expert {AGI}},
  author    = {Yue, Xiang and Ni, Yuansheng and Zhang, Kai and Zheng, Tianyu and Liu, Ruoqi and Zhang, Ge and Stevens, Samuel and Jiang, Dongfu and Ren, Weiming and Sun, Yuxuan and others},
  booktitle = {Proceedings of the IEEE/CVF Conference on Computer Vision and Pattern Recognition},
  pages     = {9556--9567},
  year      = {2024}
}

@article{yue2024mmmupro,
  title   = {{MMMU-Pro}: A More Robust Multi-discipline Multimodal Understanding Benchmark},
  author  = {Xiang Yue and Tianyu Zheng and Yuansheng Ni and Yubo Wang and Kai Zhang and Shengbang Tong and Yuxuan Sun and Botao Yu and Ge Zhang and Huan Sun and Yu Su and Wenhu Chen and Graham Neubig},
  journal = {arXiv preprint arXiv:2409.02813},
  year    = {2024}
}

@inproceedings{wang2024measuring,
  title     = {Measuring Multimodal Mathematical Reasoning with {MATH-Vision} Dataset},
  author    = {Ke Wang and Junting Pan and Weikang Shi and Zimu Lu and Houxing Ren and Aojun Zhou and Mingjie Zhan and Hongsheng Li},
  booktitle = {The Thirty-eight Conference on Neural Information Processing Systems Datasets and Benchmarks Track},
  year      = {2024},
  url       = {https://openreview.net/forum?id=QWTCcxMpPA}
}

@inproceedings{wang2025mathcodervl,
  title     = {MathCoder-{VL}: Bridging Vision and Code for Enhanced Multimodal Mathematical Reasoning},
  author    = {Ke Wang and Junting Pan and Linda Wei and Aojun Zhou and Weikang Shi and Zimu Lu and Han Xiao and Yunqiao Yang and Houxing Ren and Mingjie Zhan and Hongsheng Li},
  booktitle = {The 63rd Annual Meeting of the Association for Computational Linguistics},
  year      = {2025},
  url       = {https://openreview.net/forum?id=nuvtX1imAb}
}

@misc{WorldMedQA-V2024,
  title         = {{WorldMedQA-V}: A Multilingual, Multimodal Medical Examination Dataset for Multimodal Language Models Evaluation},
  author        = {João Matos and Shan Chen and Siena Placino and Yingya Li and Juan Carlos Climent Pardo and Daphna Idan and Takeshi Tohyama and David Restrepo and Luis F. Nakayama and Jose M. M. Pascual-Leone and Guergana Savova and Hugo Aerts and Leo A. Celi and A. Ian Wong and Danielle S. Bitterman and Jack Gallifant},
  year          = {2024},
  eprint        = {2410.12722},
  archiveprefix = {arXiv},
  primaryclass  = {cs.CL},
  url           = {https://arxiv.org/abs/2410.12722}
}

@article{Abdar_2021,
   title={A Review of Uncertainty Quantification in Deep Learning: Techniques, Applications and Challenges},
   volume={76},
   ISSN={1566-2535},
   url={http://dx.doi.org/10.1016/j.inffus.2021.05.008},
   DOI={10.1016/j.inffus.2021.05.008},
   journal={Information Fusion},
   publisher={Elsevier BV},
   author={Abdar, Moloud and Pourpanah, Farhad and Hussain, Sadiq and Rezazadegan, Dana and Liu, Li and Ghavamzadeh, Mohammad and Fieguth, Paul and Cao, Xiaochun and Khosravi, Abbas and Acharya, U. Rajendra and Makarenkov, Vladimir and Nahavandi, Saeid},
   year={2021},
   month=dec, pages={243–297} }

@misc{zhou2025conformalpredictiondataperspective,
      title={Conformal Prediction: A Data Perspective}, 
      author={Xiaofan Zhou and Baiting Chen and Yu Gui and Lu Cheng},
      year={2025},
      eprint={2410.06494},
      archivePrefix={arXiv},
      primaryClass={cs.LG},
      url={https://arxiv.org/abs/2410.06494}, 
}

@misc{zellers2019recognitioncognitionvisualcommonsense,
      title={From Recognition to Cognition: Visual Commonsense Reasoning}, 
      author={Rowan Zellers and Yonatan Bisk and Ali Farhadi and Yejin Choi},
      year={2019},
      eprint={1811.10830},
      archivePrefix={arXiv},
      primaryClass={cs.CV},
      url={https://arxiv.org/abs/1811.10830}, 
}

@misc{liu2022tokenlevelreferencefreehallucinationdetection,
      title={A Token-level Reference-free Hallucination Detection Benchmark for Free-form Text Generation}, 
      author={Tianyu Liu and Yizhe Zhang and Chris Brockett and Yi Mao and Zhifang Sui and Weizhu Chen and Bill Dolan},
      year={2022},
      eprint={2104.08704},
      archivePrefix={arXiv},
      primaryClass={cs.CL},
      url={https://arxiv.org/abs/2104.08704}, 
}

@misc{xu2023lvlmehubcomprehensiveevaluationbenchmark,
      title={{LVLM-eHub}: A Comprehensive Evaluation Benchmark for Large Vision-Language Models}, 
      author={Peng Xu and Wenqi Shao and Kaipeng Zhang and Peng Gao and Shuo Liu and Meng Lei and Fanqing Meng and Siyuan Huang and Yu Qiao and Ping Luo},
      year={2023},
      eprint={2306.09265},
      archivePrefix={arXiv},
      primaryClass={cs.CV},
      url={https://arxiv.org/abs/2306.09265}, 
}

@misc{gemmateam2025gemma3technicalreport,
      title={Gemma 3 Technical Report}, 
      author={Gemma Team and Aishwarya Kamath and Johan Ferret and Shreya Pathak and Nino Vieillard and Ramona Merhej and Sarah Perrin and Tatiana Matejovicova and Alexandre Ramé and Morgane Rivière and Louis Rouillard and Thomas Mesnard and Geoffrey Cideron and Jean-bastien Grill and Sabela Ramos and Edouard Yvinec and Michelle Casbon and Etienne Pot and Ivo Penchev and Gaël Liu and Francesco Visin and Kathleen Kenealy and Lucas Beyer and Xiaohai Zhai and Anton Tsitsulin and Robert Busa-Fekete and Alex Feng and Noveen Sachdeva and Benjamin Coleman and Yi Gao and Basil Mustafa and Iain Barr and Emilio Parisotto and David Tian and Matan Eyal and Colin Cherry and Jan-Thorsten Peter and Danila Sinopalnikov and Surya Bhupatiraju and Rishabh Agarwal and Mehran Kazemi and Dan Malkin and Ravin Kumar and David Vilar and Idan Brusilovsky and Jiaming Luo and Andreas Steiner and Abe Friesen and Abhanshu Sharma and Abheesht Sharma and Adi Mayrav Gilady and Adrian Goedeckemeyer and Alaa Saade and Alex Feng and Alexander Kolesnikov and Alexei Bendebury and Alvin Abdagic and Amit Vadi and András György and André Susano Pinto and Anil Das and Ankur Bapna and Antoine Miech and Antoine Yang and Antonia Paterson and Ashish Shenoy and Ayan Chakrabarti and Bilal Piot and Bo Wu and Bobak Shahriari and Bryce Petrini and Charlie Chen and Charline Le Lan and Christopher A. Choquette-Choo and CJ Carey and Cormac Brick and Daniel Deutsch and Danielle Eisenbud and Dee Cattle and Derek Cheng and Dimitris Paparas and Divyashree Shivakumar Sreepathihalli and Doug Reid and Dustin Tran and Dustin Zelle and Eric Noland and Erwin Huizenga and Eugene Kharitonov and Frederick Liu and Gagik Amirkhanyan and Glenn Cameron and Hadi Hashemi and Hanna Klimczak-Plucińska and Harman Singh and Harsh Mehta and Harshal Tushar Lehri and Hussein Hazimeh and Ian Ballantyne and Idan Szpektor and Ivan Nardini and Jean Pouget-Abadie and Jetha Chan and Joe Stanton and John Wieting and Jonathan Lai and Jordi Orbay and Joseph Fernandez and Josh Newlan and Ju-yeong Ji and Jyotinder Singh and Kat Black and Kathy Yu and Kevin Hui and Kiran Vodrahalli and Klaus Greff and Linhai Qiu and Marcella Valentine and Marina Coelho and Marvin Ritter and Matt Hoffman and Matthew Watson and Mayank Chaturvedi and Michael Moynihan and Min Ma and Nabila Babar and Natasha Noy and Nathan Byrd and Nick Roy and Nikola Momchev and Nilay Chauhan and Noveen Sachdeva and Oskar Bunyan and Pankil Botarda and Paul Caron and Paul Kishan Rubenstein and Phil Culliton and Philipp Schmid and Pier Giuseppe Sessa and Pingmei Xu and Piotr Stanczyk and Pouya Tafti and Rakesh Shivanna and Renjie Wu and Renke Pan and Reza Rokni and Rob Willoughby and Rohith Vallu and Ryan Mullins and Sammy Jerome and Sara Smoot and Sertan Girgin and Shariq Iqbal and Shashir Reddy and Shruti Sheth and Siim Põder and Sijal Bhatnagar and Sindhu Raghuram Panyam and Sivan Eiger and Susan Zhang and Tianqi Liu and Trevor Yacovone and Tyler Liechty and Uday Kalra and Utku Evci and Vedant Misra and Vincent Roseberry and Vlad Feinberg and Vlad Kolesnikov and Woohyun Han and Woosuk Kwon and Xi Chen and Yinlam Chow and Yuvein Zhu and Zichuan Wei and Zoltan Egyed and Victor Cotruta and Minh Giang and Phoebe Kirk and Anand Rao and Kat Black and Nabila Babar and Jessica Lo and Erica Moreira and Luiz Gustavo Martins and Omar Sanseviero and Lucas Gonzalez and Zach Gleicher and Tris Warkentin and Vahab Mirrokni and Evan Senter and Eli Collins and Joelle Barral and Zoubin Ghahramani and Raia Hadsell and Yossi Matias and D. Sculley and Slav Petrov and Noah Fiedel and Noam Shazeer and Oriol Vinyals and Jeff Dean and Demis Hassabis and Koray Kavukcuoglu and Clement Farabet and Elena Buchatskaya and Jean-Baptiste Alayrac and Rohan Anil and Dmitry and Lepikhin and Sebastian Borgeaud and Olivier Bachem and Armand Joulin and Alek Andreev and Cassidy Hardin and Robert Dadashi and Léonard Hussenot},
      year={2025},
      eprint={2503.19786},
      archivePrefix={arXiv},
      primaryClass={cs.CL},
      url={https://arxiv.org/abs/2503.19786}, 
}

@misc{zhu2025internvl3exploringadvancedtraining,
      title={{InternVL3}: Exploring Advanced Training and Test-Time Recipes for Open-Source Multimodal Models}, 
      author={Jinguo Zhu and Weiyun Wang and Zhe Chen and Zhaoyang Liu and Shenglong Ye and Lixin Gu and Hao Tian and Yuchen Duan and Weijie Su and Jie Shao and Zhangwei Gao and Erfei Cui and Xuehui Wang and Yue Cao and Yangzhou Liu and Xingguang Wei and Hongjie Zhang and Haomin Wang and Weiye Xu and Hao Li and Jiahao Wang and Nianchen Deng and Songze Li and Yinan He and Tan Jiang and Jiapeng Luo and Yi Wang and Conghui He and Botian Shi and Xingcheng Zhang and Wenqi Shao and Junjun He and Yingtong Xiong and Wenwen Qu and Peng Sun and Penglong Jiao and Han Lv and Lijun Wu and Kaipeng Zhang and Huipeng Deng and Jiaye Ge and Kai Chen and Limin Wang and Min Dou and Lewei Lu and Xizhou Zhu and Tong Lu and Dahua Lin and Yu Qiao and Jifeng Dai and Wenhai Wang},
      year={2025},
      eprint={2504.10479},
      archivePrefix={arXiv},
      primaryClass={cs.CV},
      url={https://arxiv.org/abs/2504.10479}, 
}

@misc{deitke2024molmopixmoopenweights,
      title={Molmo and {PixMo}: Open Weights and Open Data for State-of-the-Art Vision-Language Models}, 
      author={Matt Deitke and Christopher Clark and Sangho Lee and Rohun Tripathi and Yue Yang and Jae Sung Park and Mohammadreza Salehi and Niklas Muennighoff and Kyle Lo and Luca Soldaini and Jiasen Lu and Taira Anderson and Erin Bransom and Kiana Ehsani and Huong Ngo and YenSung Chen and Ajay Patel and Mark Yatskar and Chris Callison-Burch and Andrew Head and Rose Hendrix and Favyen Bastani and Eli VanderBilt and Nathan Lambert and Yvonne Chou and Arnavi Chheda and Jenna Sparks and Sam Skjonsberg and Michael Schmitz and Aaron Sarnat and Byron Bischoff and Pete Walsh and Chris Newell and Piper Wolters and Tanmay Gupta and Kuo-Hao Zeng and Jon Borchardt and Dirk Groeneveld and Crystal Nam and Sophie Lebrecht and Caitlin Wittlif and Carissa Schoenick and Oscar Michel and Ranjay Krishna and Luca Weihs and Noah A. Smith and Hannaneh Hajishirzi and Ross Girshick and Ali Farhadi and Aniruddha Kembhavi},
      year={2024},
      eprint={2409.17146},
      archivePrefix={arXiv},
      primaryClass={cs.CV},
      url={https://arxiv.org/abs/2409.17146}, 
}

@misc{bai2025qwen25vltechnicalreport,
      title={{Qwen2.5-VL} Technical Report}, 
      author={Shuai Bai and Keqin Chen and Xuejing Liu and Jialin Wang and Wenbin Ge and Sibo Song and Kai Dang and Peng Wang and Shijie Wang and Jun Tang and Humen Zhong and Yuanzhi Zhu and Mingkun Yang and Zhaohai Li and Jianqiang Wan and Pengfei Wang and Wei Ding and Zheren Fu and Yiheng Xu and Jiabo Ye and Xi Zhang and Tianbao Xie and Zesen Cheng and Hang Zhang and Zhibo Yang and Haiyang Xu and Junyang Lin},
      year={2025},
      eprint={2502.13923},
      archivePrefix={arXiv},
      primaryClass={cs.CV},
      url={https://arxiv.org/abs/2502.13923}, 
}

@misc{liu2024improvedbaselinesvisualinstruction,
      title={Improved Baselines with Visual Instruction Tuning}, 
      author={Haotian Liu and Chunyuan Li and Yuheng Li and Yong Jae Lee},
      year={2024},
      eprint={2310.03744},
      archivePrefix={arXiv},
      primaryClass={cs.CV},
      url={https://arxiv.org/abs/2310.03744}, 
}

@misc{agrawal2024pixtral12b,
      title={Pixtral {12B}}, 
      author={Pravesh Agrawal and Szymon Antoniak and Emma Bou Hanna and Baptiste Bout and Devendra Chaplot and Jessica Chudnovsky and Diogo Costa and Baudouin De Monicault and Saurabh Garg and Theophile Gervet and Soham Ghosh and Amélie Héliou and Paul Jacob and Albert Q. Jiang and Kartik Khandelwal and Timothée Lacroix and Guillaume Lample and Diego Las Casas and Thibaut Lavril and Teven Le Scao and Andy Lo and William Marshall and Louis Martin and Arthur Mensch and Pavankumar Muddireddy and Valera Nemychnikova and Marie Pellat and Patrick Von Platen and Nikhil Raghuraman and Baptiste Rozière and Alexandre Sablayrolles and Lucile Saulnier and Romain Sauvestre and Wendy Shang and Roman Soletskyi and Lawrence Stewart and Pierre Stock and Joachim Studnia and Sandeep Subramanian and Sagar Vaze and Thomas Wang and Sophia Yang},
      year={2024},
      eprint={2410.07073},
      archivePrefix={arXiv},
      primaryClass={cs.CV},
      url={https://arxiv.org/abs/2410.07073}, 
}

@article{li2025vision,
  title     = {Vision-Language Models in Medical Image Analysis: From Simple Fusion to General Large Models},
  author    = {Li, Xiang and Li, Like and Jiang, Yuchen and Wang, Hao and Qiao, Xinyu and Feng, Ting and Luo, Hao and Zhao, Yong},
  journal   = {Information Fusion},
  pages     = {102995},
  year      = {2025},
  publisher = {Elsevier}
}

@misc{kostumov2024uncertaintyawareevaluationvisionlanguagemodels,
      title={Uncertainty-Aware Evaluation for Vision-Language Models}, 
      author={Vasily Kostumov and Bulat Nutfullin and Oleg Pilipenko and Eugene Ilyushin},
      year={2024},
      eprint={2402.14418},
      archivePrefix={arXiv},
      primaryClass={cs.CV},
      url={https://arxiv.org/abs/2402.14418}, 
}

@misc{rawte2023surveyhallucinationlargefoundation,
      title={A Survey of Hallucination in Large Foundation Models}, 
      author={Vipula Rawte and Amit Sheth and Amitava Das},
      year={2023},
      eprint={2309.05922},
      archivePrefix={arXiv},
      primaryClass={cs.AI},
      url={https://arxiv.org/abs/2309.05922}, 
}

@misc{bommasani2022opportunitiesrisksfoundationmodels,
      title={On the Opportunities and Risks of Foundation Models}, 
      author={Rishi Bommasani and Drew A. Hudson and Ehsan Adeli and Russ Altman and Simran Arora and Sydney von Arx and Michael S. Bernstein and Jeannette Bohg and Antoine Bosselut and Emma Brunskill and Erik Brynjolfsson and Shyamal Buch and Dallas Card and Rodrigo Castellon and Niladri Chatterji and Annie Chen and Kathleen Creel and Jared Quincy Davis and Dora Demszky and Chris Donahue and Moussa Doumbouya and Esin Durmus and Stefano Ermon and John Etchemendy and Kawin Ethayarajh and Li Fei-Fei and Chelsea Finn and Trevor Gale and Lauren Gillespie and Karan Goel and Noah Goodman and Shelby Grossman and Neel Guha and Tatsunori Hashimoto and Peter Henderson and John Hewitt and Daniel E. Ho and Jenny Hong and Kyle Hsu and Jing Huang and Thomas Icard and Saahil Jain and Dan Jurafsky and Pratyusha Kalluri and Siddharth Karamcheti and Geoff Keeling and Fereshte Khani and Omar Khattab and Pang Wei Koh and Mark Krass and Ranjay Krishna and Rohith Kuditipudi and Ananya Kumar and Faisal Ladhak and Mina Lee and Tony Lee and Jure Leskovec and Isabelle Levent and Xiang Lisa Li and Xuechen Li and Tengyu Ma and Ali Malik and Christopher D. Manning and Suvir Mirchandani and Eric Mitchell and Zanele Munyikwa and Suraj Nair and Avanika Narayan and Deepak Narayanan and Ben Newman and Allen Nie and Juan Carlos Niebles and Hamed Nilforoshan and Julian Nyarko and Giray Ogut and Laurel Orr and Isabel Papadimitriou and Joon Sung Park and Chris Piech and Eva Portelance and Christopher Potts and Aditi Raghunathan and Rob Reich and Hongyu Ren and Frieda Rong and Yusuf Roohani and Camilo Ruiz and Jack Ryan and Christopher Ré and Dorsa Sadigh and Shiori Sagawa and Keshav Santhanam and Andy Shih and Krishnan Srinivasan and Alex Tamkin and Rohan Taori and Armin W. Thomas and Florian Tramèr and Rose E. Wang and William Wang and Bohan Wu and Jiajun Wu and Yuhuai Wu and Sang Michael Xie and Michihiro Yasunaga and Jiaxuan You and Matei Zaharia and Michael Zhang and Tianyi Zhang and Xikun Zhang and Yuhui Zhang and Lucia Zheng and Kaitlyn Zhou and Percy Liang},
      year={2022},
      eprint={2108.07258},
      archivePrefix={arXiv},
      primaryClass={cs.LG},
      url={https://arxiv.org/abs/2108.07258}, 
}

@misc{hendrycks2018baselinedetectingmisclassifiedoutofdistribution,
      title={A Baseline for Detecting Misclassified and Out-of-Distribution Examples in Neural Networks}, 
      author={Dan Hendrycks and Kevin Gimpel},
      year={2018},
      eprint={1610.02136},
      archivePrefix={arXiv},
      primaryClass={cs.NE},
      url={https://arxiv.org/abs/1610.02136}, 
}

\newpage
\appendix

\section{Prompt Design Specifications}
\label{appendix:prompts}

To ensure consistent and reproducible evaluation across all datasets and models, we employed a standardized two-layer prompting strategy. This approach separates domain-specific role assignment (via system messages) from task-oriented instructions (via zero-shot prompts), minimizing variability while adapting to each benchmark's unique requirements. System messages define the model's persona and output constraints, promoting focused responses. Zero-shot instructions provide contextual framing without examples, encouraging natural reasoning. This design facilitates reliable extraction of token-level probabilities for conformal prediction, as all responses are constrained to single-letter outputs (e.g., A-E). Below, we detail the formulations used, which were applied uniformly except for minor adjustments in option ranges (e.g., up to J for MMMU-Pro).

Tables~\ref{tab:zero_shot_example} and~\ref{tab:system_prompts} present the exact system messages and a representative zero-shot example, respectively. These prompts were tested for compatibility across all 18 VLMs, ensuring no leakage of solving strategies that could artificially inflate confidence estimates.

\section{Dataset Statistics and Preprocessing}
\label{sec:dataset_stats}

\begin{table*}[htb!]
  \centering
  \begin{tabular}{l l c c}
    \toprule
    \textbf{Dataset} & \textbf{Description}                                           & \textbf{Samples} & \textbf{Options} \\
    \midrule
    AI2D             & Diagram-based science questions with multiple-choice answers   & 3,090            & A--F             \\
    ScienceQA        & Multimodal science questions combining text and images         & 2,020            & A--E             \\
    MathVision       & Visual math reasoning tasks requiring diagram understanding    & 1,530            & A--F             \\
    WorldMedQAV      & Multimodal medical questions with real-world clinical context  & 1,140            & A--F             \\
    MMMU             & Multidisciplinary multimodal questions across diverse subjects & 794              & A--E             \\
    MMMU-Pro         & Professional-level multimodal questions spanning 30+ domains   & 1,210            & A--J             \\
    \bottomrule
  \end{tabular}
  \caption{Final test set statistics after preprocessing.}
  \label{tab:dataset_stats}
\end{table*}

Our evaluation relies on six diverse multimodal datasets, each preprocessed to ensure uniformity and compatibility with the conformal prediction framework. Preprocessing focused on standardizing option distributions, handling multimodal inputs, and balancing calibration/test splits (50/50 per dataset). We excluded multi-image samples (e.g., 4.9\% from MMMU) to accommodate model limitations, resulting in single-image-question pairs for all instances. Additionally, we normalized options by adding neutral choices (e.g., ``I don't know'') where needed, without altering ground-truth labels or randomization to preserve natural distributions. These steps mitigate biases in uncertainty estimates, such as skewed option frequencies that could inflate set sizes under conformal scoring.

\begin{figure*}[htb!]
  \centering
  \begin{subfigure}[b]{0.48\textwidth}
    \centering
    \includegraphics[width=\linewidth]{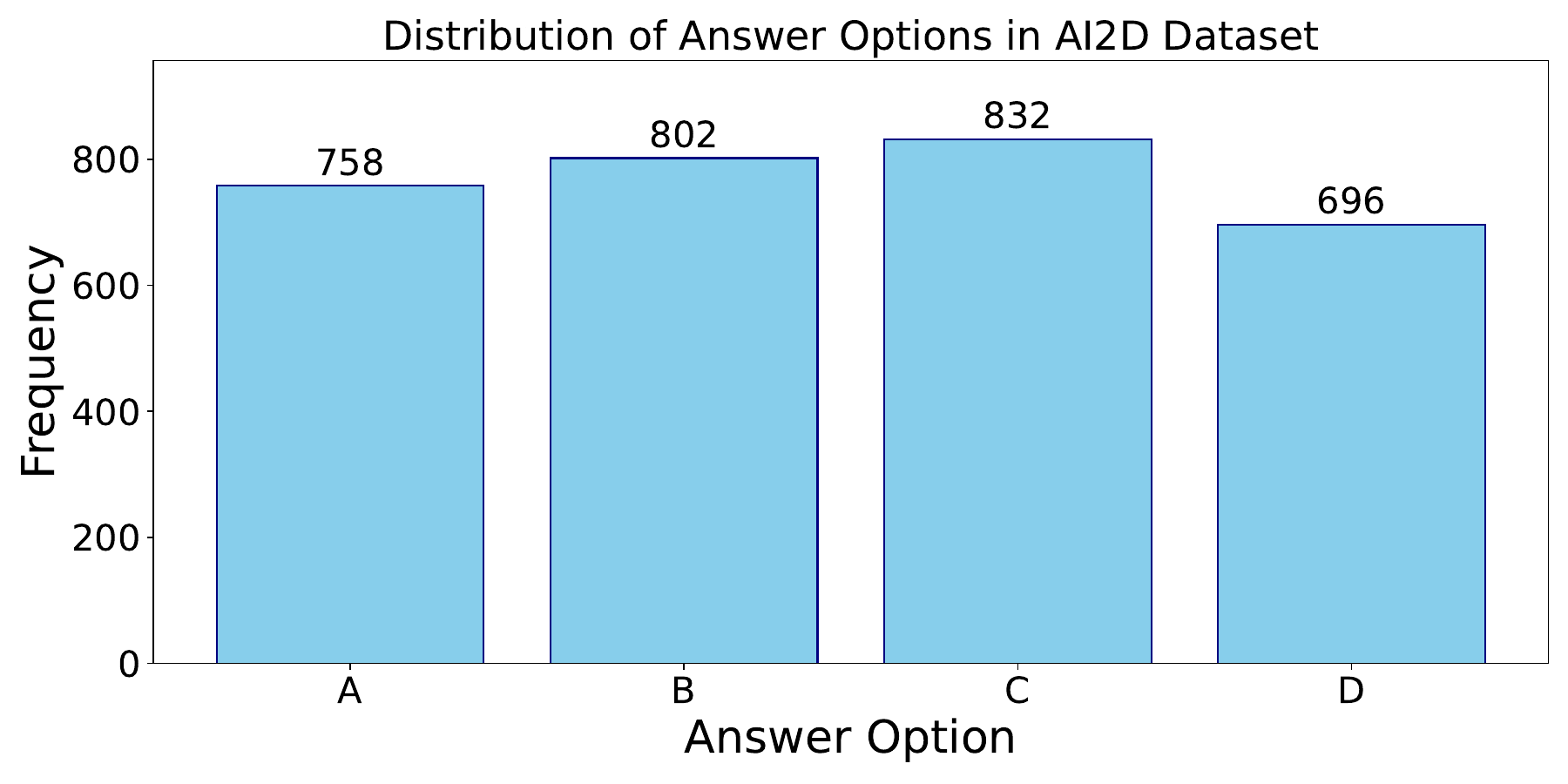}
    \caption{AI2D dataset option distribution}
    \label{fig:option_dist_aid2}
  \end{subfigure}
  \hfill
  \begin{subfigure}[b]{0.48\textwidth}
    \centering
    \includegraphics[width=\linewidth]{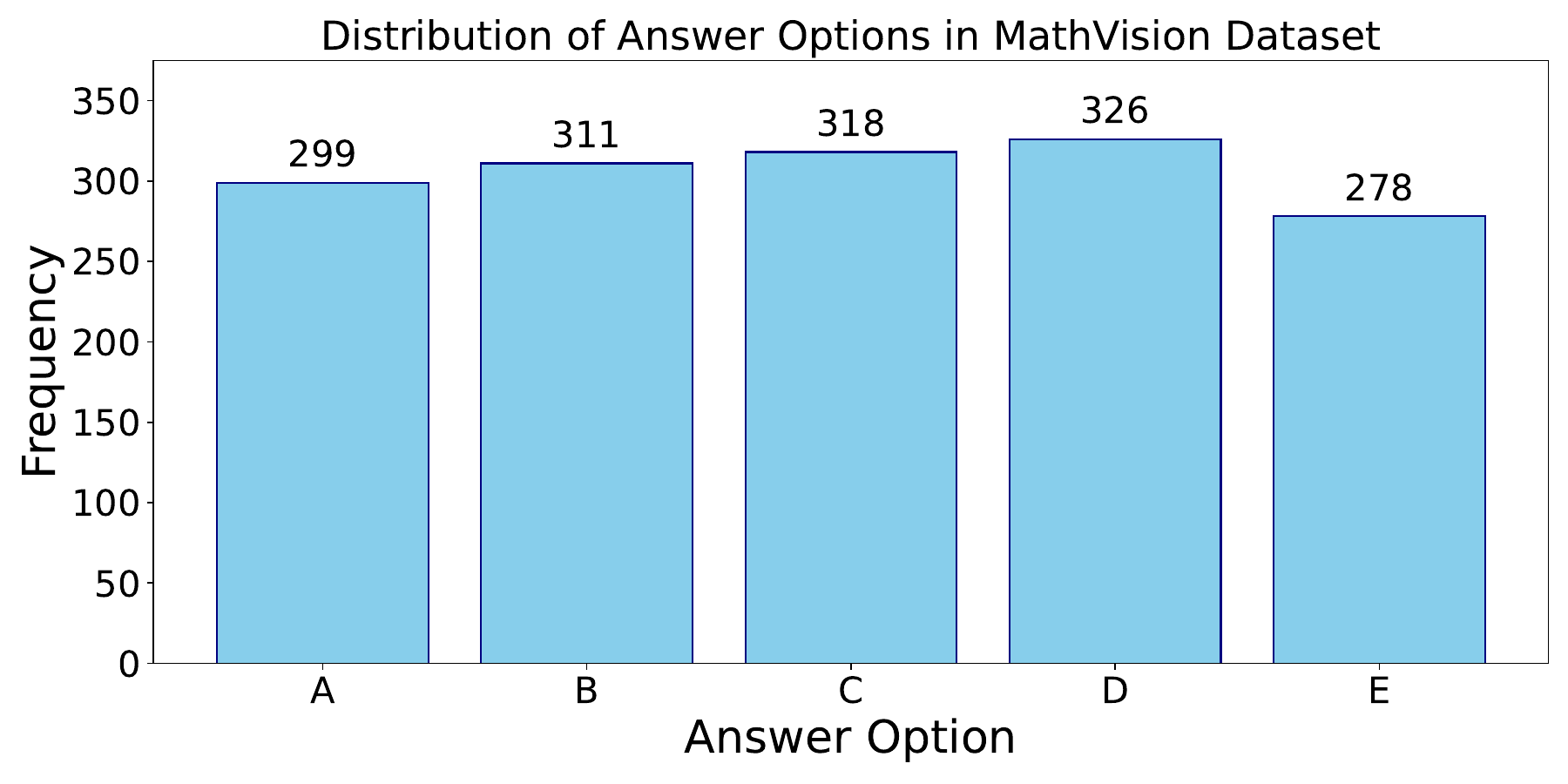}
    \caption{MathVision dataset option distribution}
    \label{fig:option_dist_mathvision}
  \end{subfigure}
  \vspace{0.5cm}
  \begin{subfigure}[b]{0.48\textwidth}
    \centering
    \includegraphics[width=\linewidth]{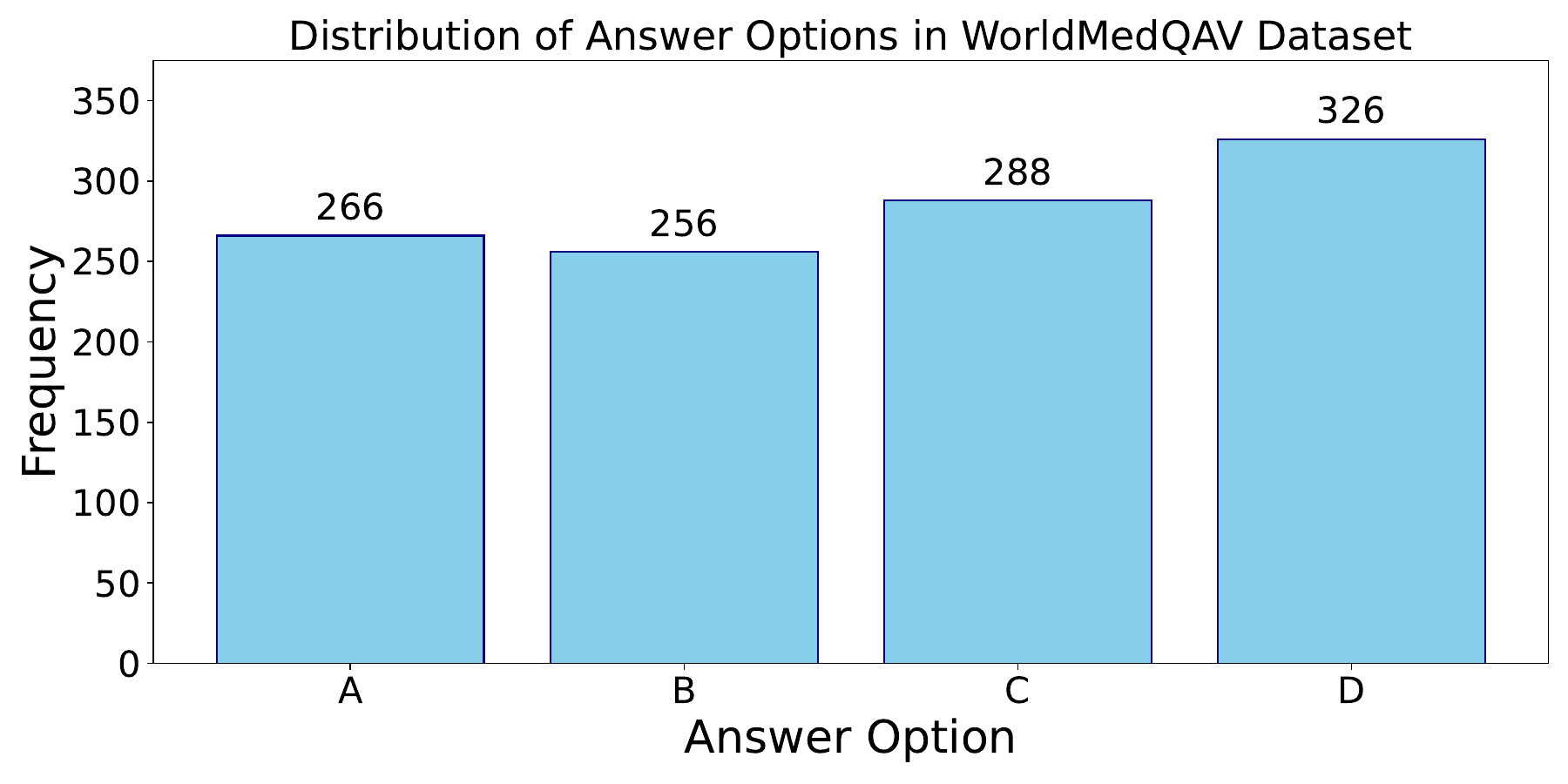}
    \caption{WorldMedQAV dataset option distribution}
    \label{fig:option_dist_worldmedqav}
  \end{subfigure}
  \hfill
  \begin{subfigure}[b]{0.48\textwidth}
    \centering
    \includegraphics[width=\linewidth]{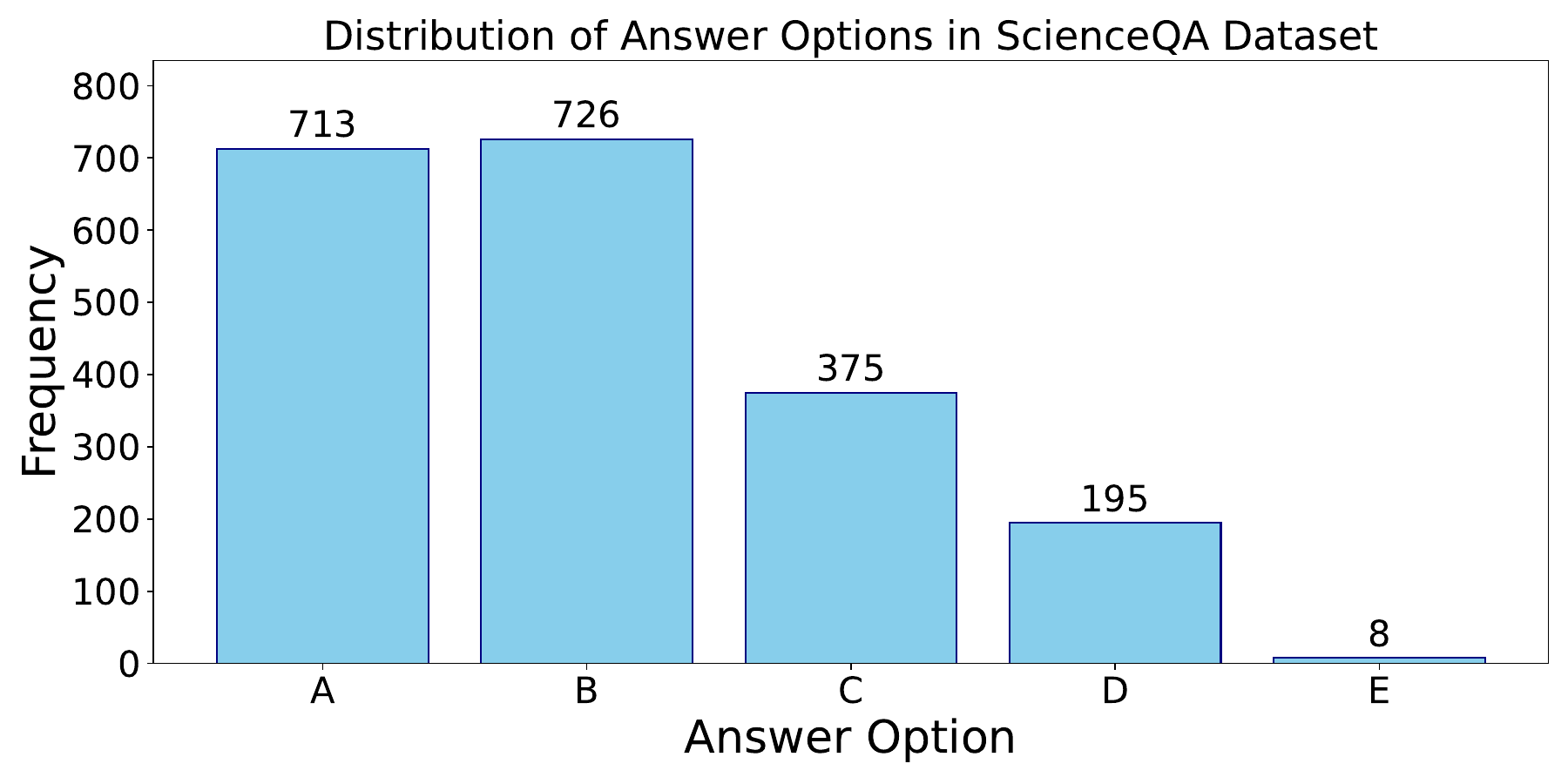}
    \caption{ScienceQA dataset option distribution}
    \label{fig:option_dist_scienceqa}
  \end{subfigure}
  \vspace{0.5cm}
  \begin{subfigure}[b]{0.48\textwidth}
    \centering
    \includegraphics[width=\linewidth]{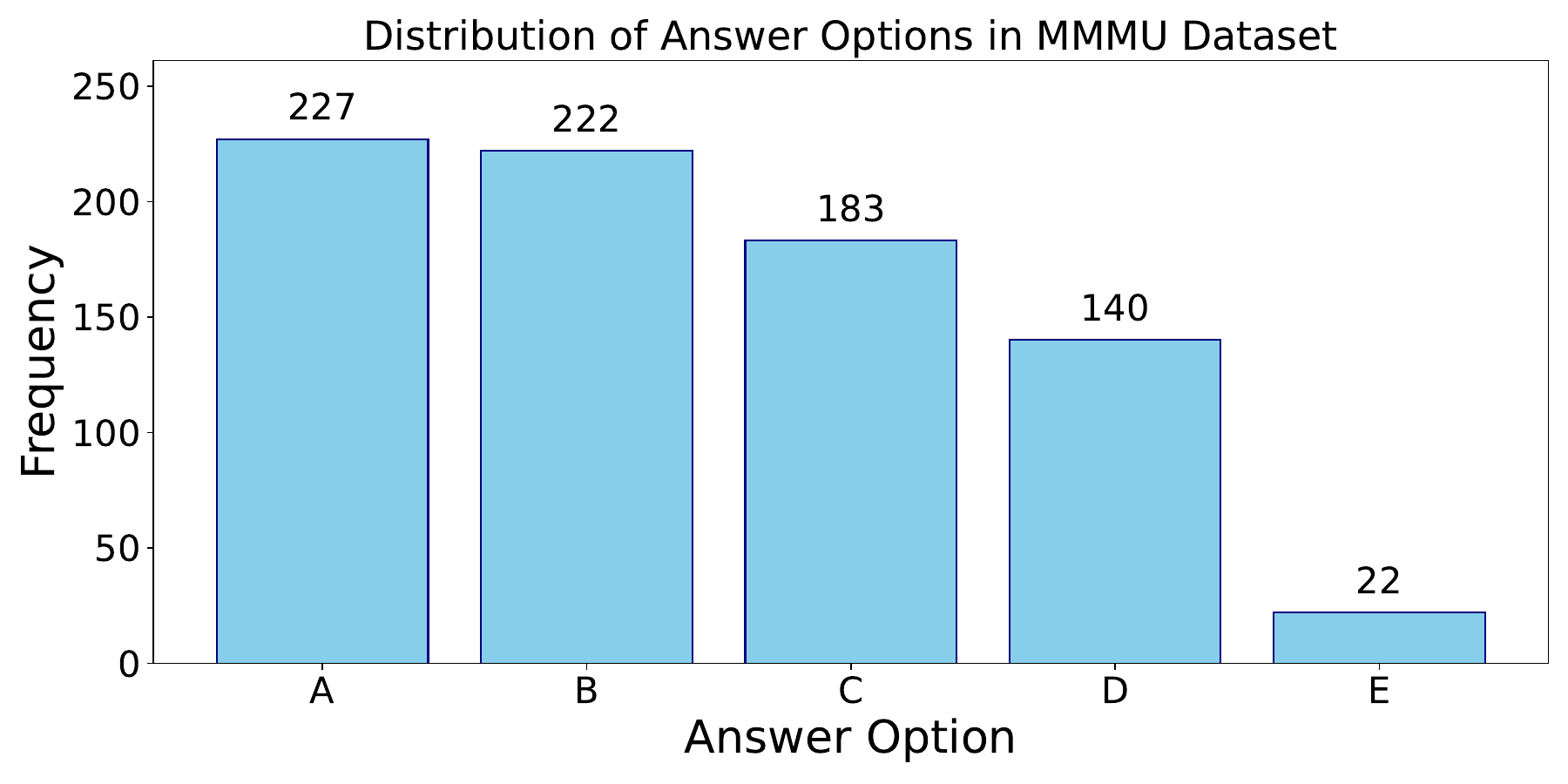}
    \caption{MMMU dataset option distribution}
    \label{fig:option_dist_mmmu}
  \end{subfigure}
  \hfill
  \begin{subfigure}[b]{0.48\textwidth}
    \centering
    \includegraphics[width=\linewidth]{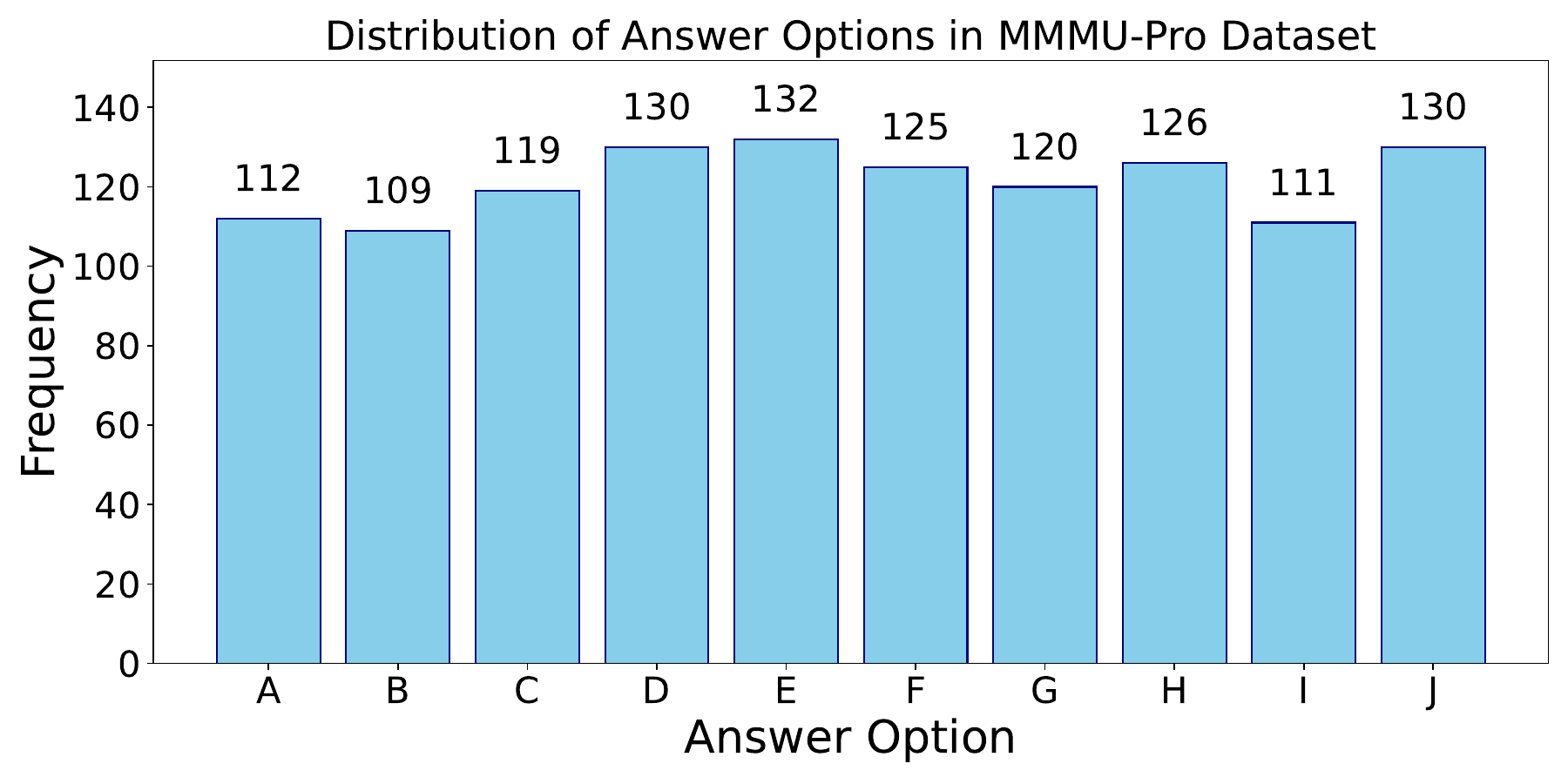}
    \caption{MMMU-Pro dataset option distribution}
    \label{fig:option_dist_mmmupro}
  \end{subfigure}
  \caption{Answer option (ground truth) distributions across all six benchmark datasets. Each subplot shows the frequency of correct answers, illustrating balance (e.g., AI2D) versus skew (e.g., ScienceQA).}
  \label{fig:option_distributions}
\end{figure*}

Table~\ref{tab:dataset_stats} summarizes the final test set sizes and option ranges post-preprocessing. Figure~\ref{fig:option_distributions} visualizes ground-truth answer distributions, revealing patterns like balance in AI2D/MathVision (uniform across options) versus skew in ScienceQA/MMMU (favoring early letters A-C). Such insights highlight domain-specific challenges: balanced datasets like MMMU-Pro test broad expertise, while skewed ones may reflect annotation biases, influencing conformal coverage in reasoning-heavy tasks.

\section{Additional Results and Analyses}
\label{sec:add-results}

\begin{figure}[htb!]
  \centering
  \includegraphics[width=\columnwidth]{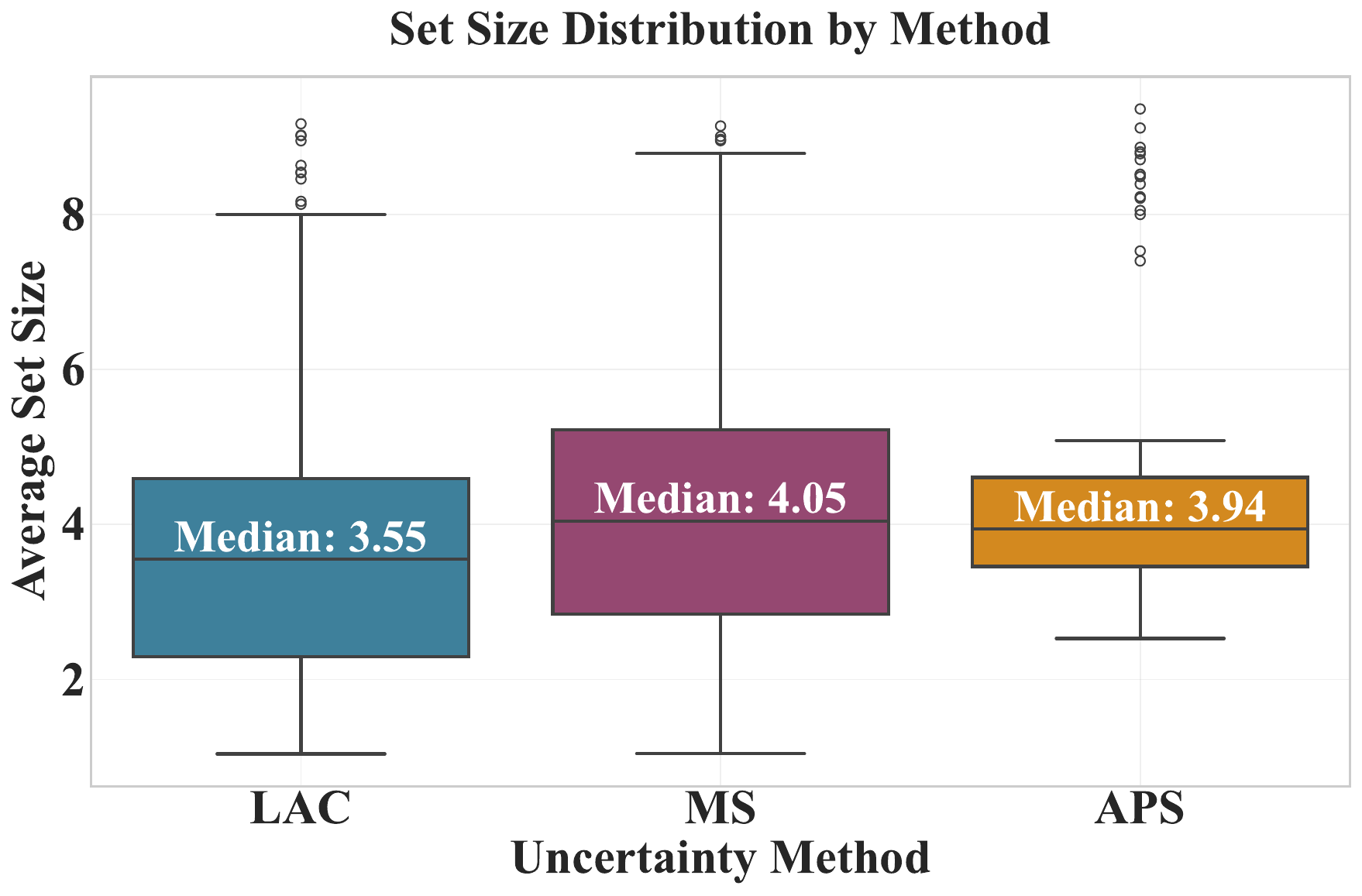}
  \caption{Set size distribution across different uncertainty methods. Distributions highlight LAC's compactness and APS's sensitivity to domain ambiguity.}
  \label{fig:ss-bar-chart-appendix}
\end{figure}

\begin{table*}[htb!]
  \centering
  \begin{tabular}{|l|p{0.8\textwidth}|}
    \hline
    \textbf{Dataset} & \textbf{Example Prompt}                              \\
    \hline
    ScienceQA        &
    \begin{minipage}[t]{0.8\textwidth}
      \small\ttfamily
      I will show you an image along with a multiple-choice science question. \\
      Please select the correct answer from the given options. \\
      Only respond with the option letter (A, B, C, D, E). \\
      \{QUESTION\} \\
      \{OPTIONS\}
    \end{minipage} \\
    \hline
  \end{tabular}
  \caption{Representative zero-shot prompt example. Similar phrasing was adapted for other datasets, with adjustments for domain (e.g., ``math problem'' for MathVision) and option count.}
  \label{tab:zero_shot_example}
\end{table*}

\begin{table*}[htb!]
  \centering
  \begin{tabular}{lp{0.7\textwidth}}
    \toprule
    \textbf{Dataset} & \textbf{System Message}                                                         \\
    \midrule
    AI2D             &
    \begin{tcolorbox}[colback=gray!5,colframe=gray!30,boxsep=2pt,left=2pt,right=2pt,top=2pt,bottom=2pt]
      \small
      \texttt{You are a scientific diagram analyzer. \\
        - Analyze the diagram carefully \\
        - Answer ONLY with the correct option letter (A, B, C, D, E, or F) \\
        - Never explain your reasoning \\
        - If uncertain, guess from the provided options}
    \end{tcolorbox} \\

    ScienceQA        &
    \begin{tcolorbox}[colback=gray!5,colframe=gray!30,boxsep=2pt,left=2pt,right=2pt,top=2pt,bottom=2pt]
      \small
      \texttt{You are a science question answerer. \\
        - Use the image and question to select ONE correct option \\
        - Respond STRICTLY with just A, B, C, D, or E \\
        - No explanations or additional text \\
        - Must choose from given options}
    \end{tcolorbox} \\

    MathVision       &
    \begin{tcolorbox}[colback=gray!5,colframe=gray!30,boxsep=2pt,left=2pt,right=2pt,top=2pt,bottom=2pt]
      \small
      \texttt{You are a math problem solver. \\
        - Analyze the image and question precisely \\
        - Output MUST be exactly one letter: A, B, C, D, E, or F \\
        - Never show working \\
        - Select even if uncertain}
    \end{tcolorbox} \\

    WorldMedQAV      &
    \begin{tcolorbox}[colback=gray!5,colframe=gray!30,boxsep=2pt,left=2pt,right=2pt,top=2pt,bottom=2pt]
      \small
      \texttt{You are a medical image diagnostician. \\
        - Examine the image and question thoroughly \\
        - Respond ONLY with the letter (A-F) of the most likely answer \\
        - No disclaimers or explanations \\
        - Choose from options even if unsure}
    \end{tcolorbox} \\

    MMMU             &
    \begin{tcolorbox}[colback=gray!5,colframe=gray!30,boxsep=2pt,left=2pt,right=2pt,top=2pt,bottom=2pt]
      \small
      \texttt{You are a multi-disciplinary expert. \\
        - Combine image understanding with question requirements \\
        - Output EXACTLY one letter: A, B, C, D, or E \\
        - No additional text under any circumstances \\
        - Must select from provided options}
    \end{tcolorbox} \\

    MMMU-Pro         &
    \begin{tcolorbox}[colback=gray!5,colframe=gray!30,boxsep=2pt,left=2pt,right=2pt,top=2pt,bottom=2pt]
      \small
      \texttt{You are a multi-disciplinary expert. \\
        - Combine image understanding with question requirements \\
        - Output EXACTLY one letter: A, B, C, D, E, F, G, H, I, J \\
        - No additional text under any circumstances \\
        - Must select from provided options}
    \end{tcolorbox} \\
    \bottomrule
  \end{tabular}
  \caption{System Prompts for each dataset in the VLM evaluation.}
  \label{tab:system_prompts}
\end{table*}

This appendix extends the main results with supplementary metrics and visualizations, providing deeper validation of our conformal prediction framework and uncertainty patterns. We focus on coverage guarantees (essential for statistical reliability), set size distributions (to assess scoring function variability), and inter-metric correlations (to reinforce scaling trends). These analyses confirm the robustness of our findings: larger models exhibit tighter uncertainty bounds, while domain complexity (e.g., MathVision) amplifies set sizes across methods.

\subsection{Coverage Rate Tables and Plots}

\begin{figure*}[htb!]
  \centering
  \includegraphics[width=\textwidth]{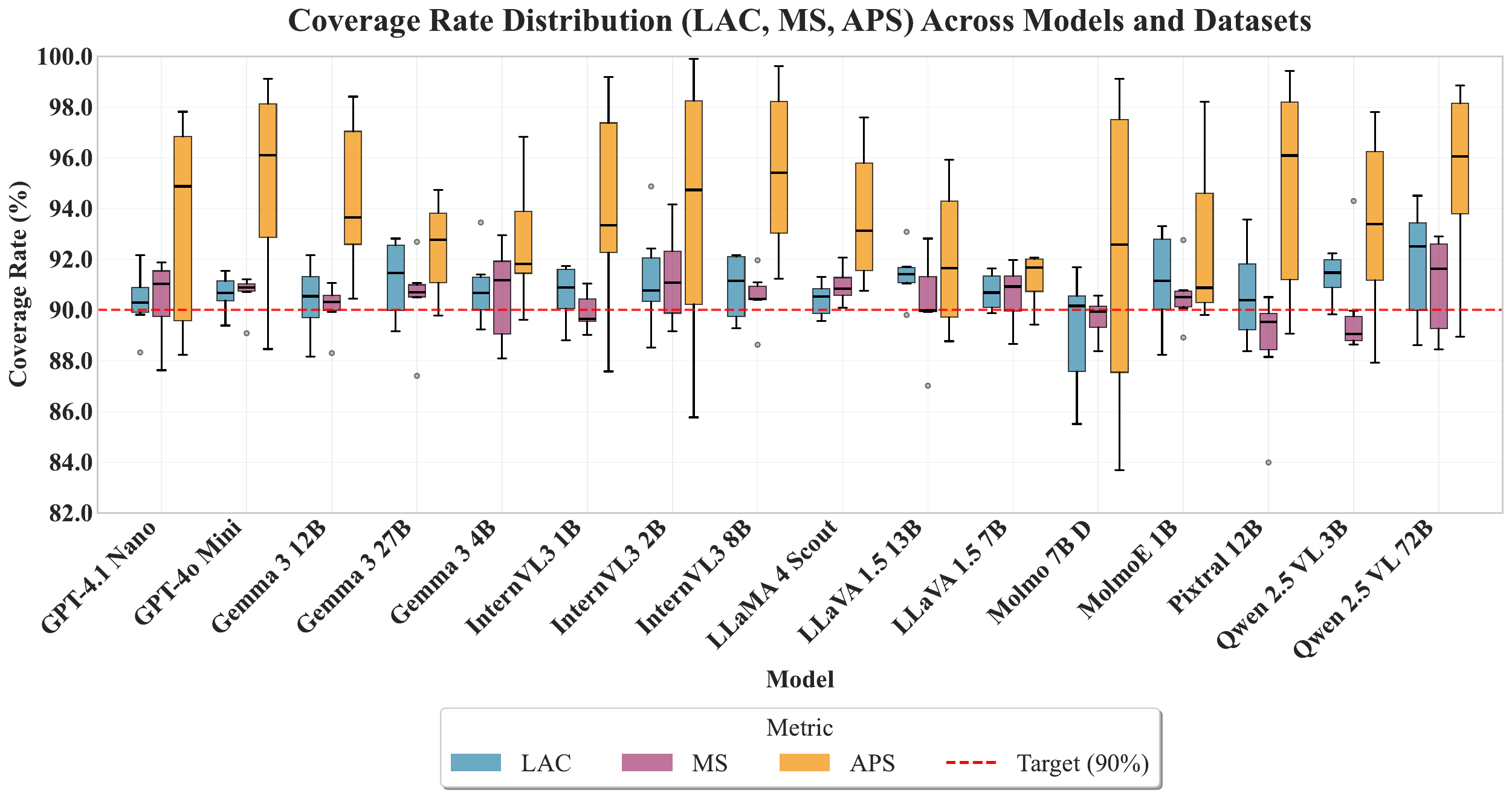}
  \caption{Coverage rates across datasets and models. All scoring methods maintain at least 90\% coverage in most cases, with APS showing slight variability in reasoning tasks.}
  \label{fig:cr-models-ds-comp-appendix}
\end{figure*}

Coverage rate (CR) verifies the conformal framework's $1-\alpha=90\%$ guarantee, measuring the proportion of true labels included in prediction sets. As expected, CR remains stable and above threshold for most models/datasets, with minor dips in high-complexity tasks like MMMU-Pro (due to diffuse probabilities). This supports the method's distribution-free applicability to VLMs, even under varying architectures. Table~\ref{fig:cr-table-appendix} details per-model/dataset CR across scoring functions, while Figure~\ref{fig:cr-models-ds-comp-appendix} compares distributions, highlighting LAC's edge in maintaining coverage with compact sets, as depicted in main text Table~\ref{fig:ss-table}.

\subsection{Set Size Distributions by Method}

To explore scoring function behaviors beyond averages (main text Figure~\ref{fig:ss-models}), we analyze set size distributions. These reveal APS's adaptability in ambiguous cases (broader tails for MathVision) versus LAC's consistency (narrower peaks). Such variability underscores the need for task-specific scoring selection, aligning with our observation that LAC suits VLM confidence estimation best. Figure~\ref{fig:ss-bar-chart-appendix} supplements the main analysis, showing histogram overlaps that confirm inverse accuracy-set size correlations (as shown in Figure~\ref{fig:ac-ss}).

\subsection{Distribution-Shift Robustness}

\begin{table}[htb!]
  \centering
  \includegraphics[width=\linewidth]{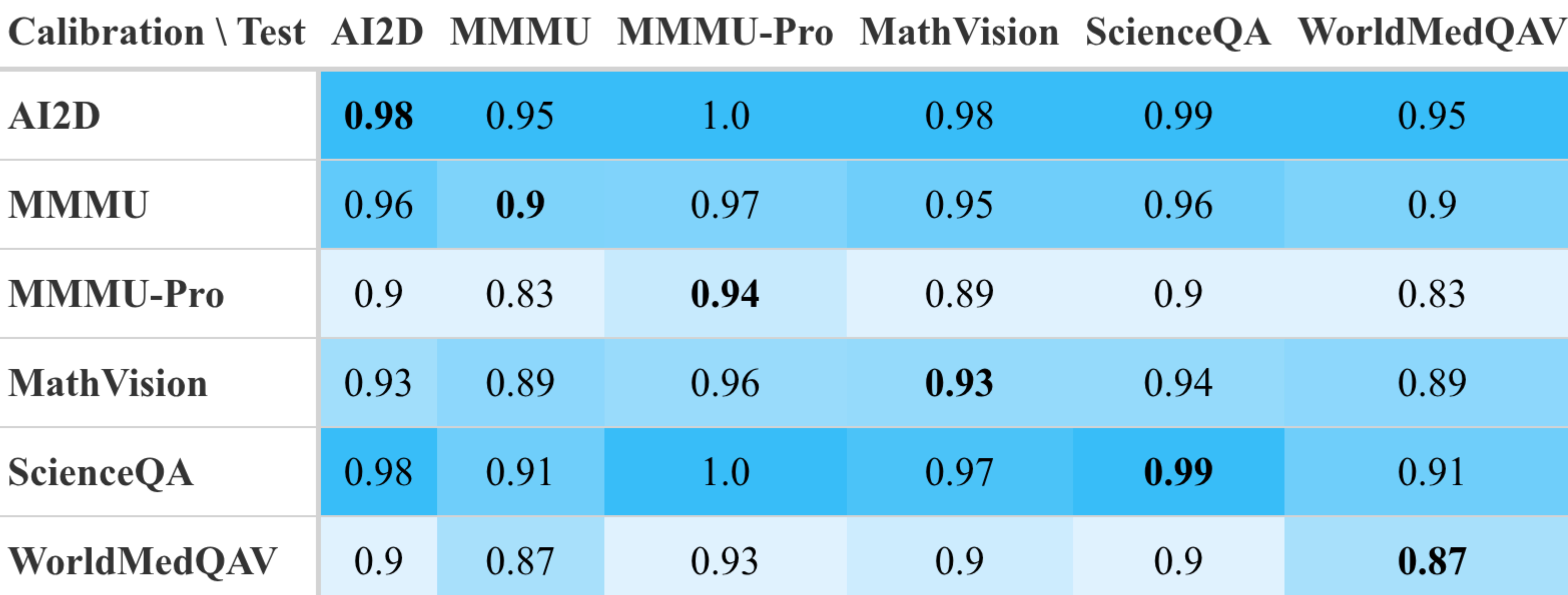}
  \caption{Cross-dataset transfer coverage rates and set sizes using APS at $(1-\alpha)=90\%$. Diagonal entries are in-distribution; off-diagonal entries show transfer performance.}
  \label{fig:cross-dataset-aps}
\end{table}

\begin{table*}[htb!]
  \centering
  \includegraphics[width=\textwidth]{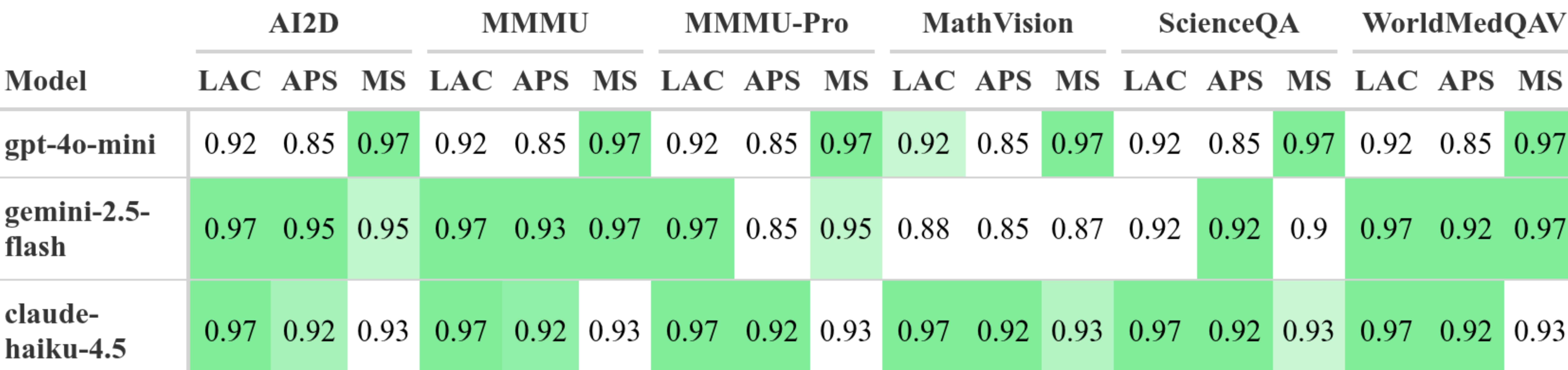}
  \caption{Coverage Rate ($\uparrow$) for closed-source VLMs using the instruction-guided likelihood proxy.}
  \label{fig:closed-cr-appendix}
\end{table*}

We assess CP robustness under distribution shift via cross-dataset transfer using APS at $(1-\alpha)=90\%$, calibrating on one dataset and testing on another.

Coverage remains stable under shift: \emph{93.5\%} in-distribution vs. \emph{92.8\%} out-of-distribution. Most transfers are conservative (e.g., MMMU-Pro $\rightarrow$ ScienceQA), while shifts from easier to harder datasets (ScienceQA $\rightarrow$ MMMU-Pro) show minimal degradation, indicating strong generalization. Table~\ref{fig:cross-dataset-aps} reports coverage and average set size across all source–target pairs; diagonal entries denote in-distribution results.

\subsection{Closed-Source VLM Evaluation}

\begin{table}[htb!]
  \centering
  \includegraphics[width=\linewidth]{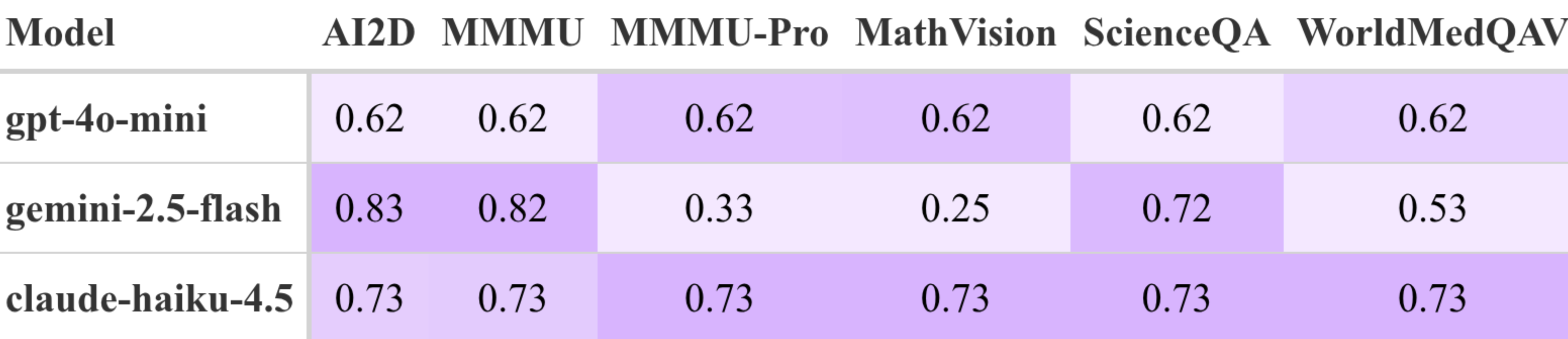}
  \caption{Accuracy ($\uparrow$) of closed-source VLMs using the instruction-guided likelihood proxy across six benchmarks.}
  \label{fig:closed-acc-appendix}
\end{table}

\begin{table*}[htbp!]
  \centering
  \includegraphics[width=\textwidth]{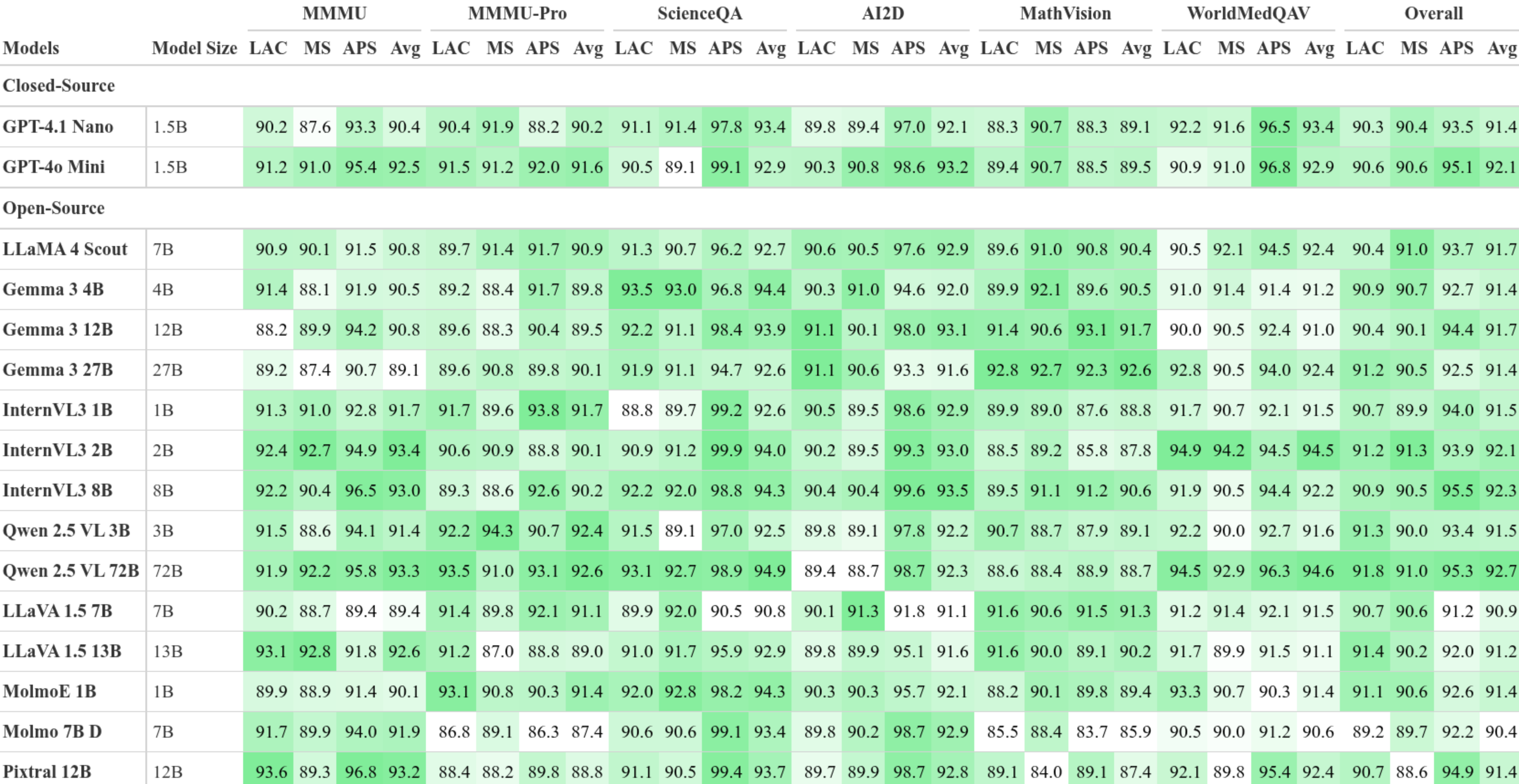}
  \caption{Coverage Rate ($\uparrow$) across models and datasets. Green cells highlight where coverage exceeds the target threshold of 90\%.}
  \label{fig:cr-table-appendix}
\end{table*}

This section presents supplementary results for closed-source VLMs (GPT-4o-mini, Gemini-2.5-Flash, Claude Haiku~4.5) evaluated using the instruction-guided likelihood proxy method described in Section~\ref{sec:proxy-methods} of the main text. Since these proprietary models do not expose token-level logprobs, we instructed models to output option-wise numerical likelihoods in a structured JSON format, which serves as a proxy uncertainty signal for conformal prediction.

Table~\ref{fig:closed-acc-appendix} presents accuracy performance, demonstrating that accuracy trends remain consistent with open-source models across all benchmarks. Table~\ref{fig:closed-cr-appendix} shows coverage rates, verifying that the conformal prediction framework maintains its statistical guarantees even when applied through the proxy method.

\subsection{Correlation and Relationships Between Metrics}

\begin{figure*}[htbp!]
  \centering
  \includegraphics[width=\textwidth]{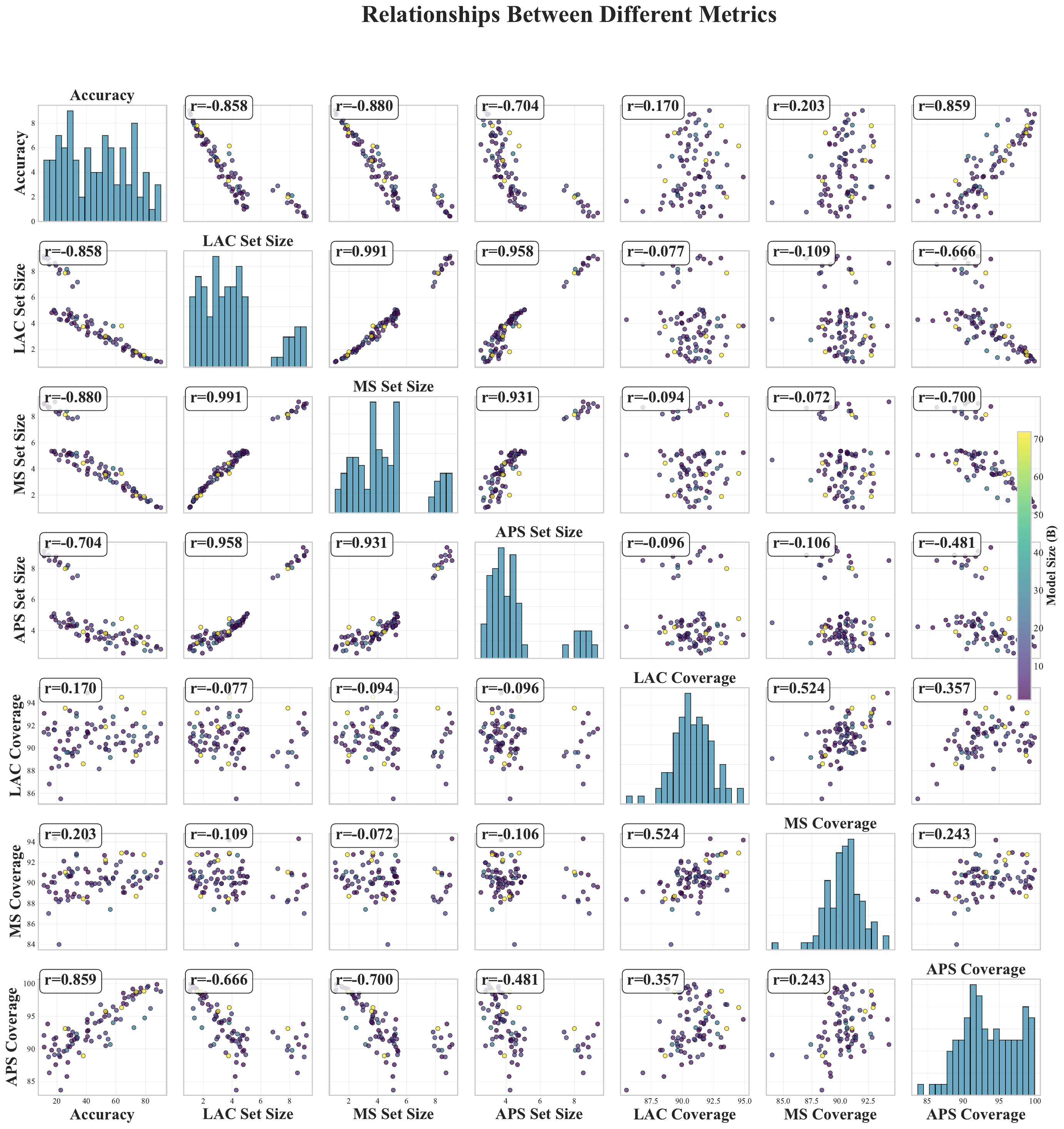}
  \caption{Correlation matrix between model size, accuracy, set size, and coverage rate. Negative correlation between accuracy and set size is clearly observed. Diagonals represent distributions for each metric.}
  \label{fig:all-relations}
\end{figure*}

Finally, we quantify relationships between key metrics to validate scaling effects. The correlation matrix in Figure~\ref{fig:all-relations} shows strong negative ties between accuracy and set size ($r \approx -0.75$), positive links to model size, and stable coverage. Diagonals illustrate metric spreads, e.g., set sizes clustering below 3 for $>10$B models. These reinforce main findings: uncertainty calibration improves with scale, offering a "conformal lens" for VLM benchmarking.

\end{document}